\documentclass[lettersize,journal]{IEEEtran}
\usepackage{amsmath,amsfonts}
\usepackage{algorithmic}
\usepackage{algorithm}
\usepackage{array}
\usepackage[caption=false,font=normalsize,labelfont=sf,textfont=sf]{subfig}
\usepackage{textcomp}
\usepackage{stfloats}
\usepackage{url}
\usepackage{verbatim}
\usepackage{graphicx}
\usepackage{cite}

\newtheorem{remark}{Remark}
\newtheorem{definition}{Definition}

\usepackage{booktabs}
\usepackage{tabularx}
\usepackage{threeparttable}
\usepackage[colorlinks,bookmarksopen=true, bookmarksnumbered=true]{hyperref}
\usepackage{multirow}

\usepackage{ragged2e}

\usepackage[edges]{forest} 
\usepackage{booktabs,tabularx}
\usepackage{amssymb}
\usepackage{placeins}

\usepackage[dvipsnames,svgnames,table]{xcolor}
\usepackage{forest}
\usepackage{ragged2e}

\ifCLASSOPTIONcompsoc
  \usepackage[nocompress]{cite}
\else
  \usepackage{cite}
\fi

\ifCLASSINFOpdf
\else
\fi

\begin{document}

\title{Out-of-Distribution Detection on Graphs:\\A Survey}
\author{Tingyi Cai,
Yunliang Jiang,~\IEEEmembership{Senior Member,~IEEE,}
Yixin Liu,
Zhonglong Zheng,~\IEEEmembership{Member,~IEEE,}\\
Changqin Huang,~\IEEEmembership{Member,~IEEE,}
Ming Li,~\IEEEmembership{Member,~IEEE,}
Shirui Pan,~\IEEEmembership{Senior Member,~IEEE}
\thanks{Tingyi Cai is with the Zhejiang Key Laboratory of Intelligent Education Technology and Application, Zhejiang Normal University, Jinhua, China,
and also with the School of Physical and Mathematical Sciences, Nanyang Technological University, Singapore, Singapore.
E-mail: tingyicai@zjnu.edu.cn.\\
Yunliang Jiang is with the Zhejiang Key Laboratory of Intelligent Education Technology and Application, Zhejiang Normal University, Jinhua, China,
and also with the School of Artificial Intelligence, Huzhou Normal University, Huzhou, China.
E-mail: jyl2022@zjnu.cn.\\
Yixin Liu is with the School of Information and Communication Technology, Griffith University, Queensland, Australia.
E-mail: yixin.liu@griffith.edu.au.\\
Zhonglong Zheng is with the School of Computer Science and Technology, Zhejiang Normal University, Jinhua, China.
E-mail: zhonglong@zjnu.edu.cn.\\
Changqin Huang is with Zhejiang University, Hangzhou, China.
E-mail: cqhuang@zju.edu.cn.\\
Ming Li is with the Zhejiang Key Laboratory of Intelligent Education Technology and Application, Zhejiang Normal University, Jinhua, China.
E-mail: mingli@zjnu.edu.cn.\\
Shirui Pan is with the School of Information and Communication Technology, Griffith University, Queensland, Australia.
E-mail: s.pan@griffith.edu.au.}
}


\markboth{Journal of \LaTeX\ Class Files,~Vol.~14, No.~8, August~2021}%
{Cai \MakeLowercase{\textit{et al.}}: Out-of-Distribution Detection on Graphs: A Survey}

\IEEEtitleabstractindextext{%
\begin{abstract}
\justifying
    Graph learning has advanced prediction and discovery across a wide range of relational data applications.
    In evolving environments, new categories and changes in attributes or relationships can produce out-of-distribution (OOD) inputs on which models make confident but unreliable predictions.
    Graph out-of-distribution (GOOD) detection identifies such nodes or graphs to support rejection and further assessment.
    This paper presents the first systematic survey dedicated to GOOD detection. We clarify the problem settings and organize existing methods into six categories according to their primary OOD discrimination mechanisms. The review examines their technical connections, supervision and reference requirements, and conditions of use, alongside representative datasets, OOD constructions, benchmarks, and evaluation metrics.
    Controlled synthetic comparisons deepen the understanding of detection settings by illustrating how representative methods respond to different distribution shifts and graph conditions.
    We discuss open challenges and outline future research directions, aiming to provide a common foundation for advancing GOOD detection.
\end{abstract}

}
\maketitle
\IEEEdisplaynontitleabstractindextext
\IEEEpeerreviewmaketitle


\section{Introduction}
\label{sec:introduction}

\IEEEPARstart{G}{raphs} describe relationships in transaction
networks, biological systems, and molecular
structures~\cite{wu2021gnnsurvey,hu2020ogb}. Their widespread use
has motivated graph learning methods for prediction, including
graph kernels, message-passing networks, and graph
transformers~\cite{shervashidze2011wl,ying2021graphormer,rampasek2022graphgps,wu2022nodeformer}.
Message-passing graph neural networks (GNNs) update node
representations through neighborhood aggregation, which
combines information from neighboring nodes~\cite{gilmer2017mpnn}.
In conventional supervised learning, training data are expected
to represent deployment inputs~\cite{wu2023gnnsafe}.
However, the input distribution can change as graph data evolve
or models are deployed in new
domains~\cite{hu2020ogb,zhang2022efraudcom,wu2018moleculenet}.
Models may produce confident but unreliable predictions on such
out-of-distribution (OOD)
inputs~\cite{ovadia2019trust,wu2023gnnsafe}.
These predictions can put downstream decisions at risk.

\begin{figure}[t]
    \centering
    \includegraphics[width=\linewidth]{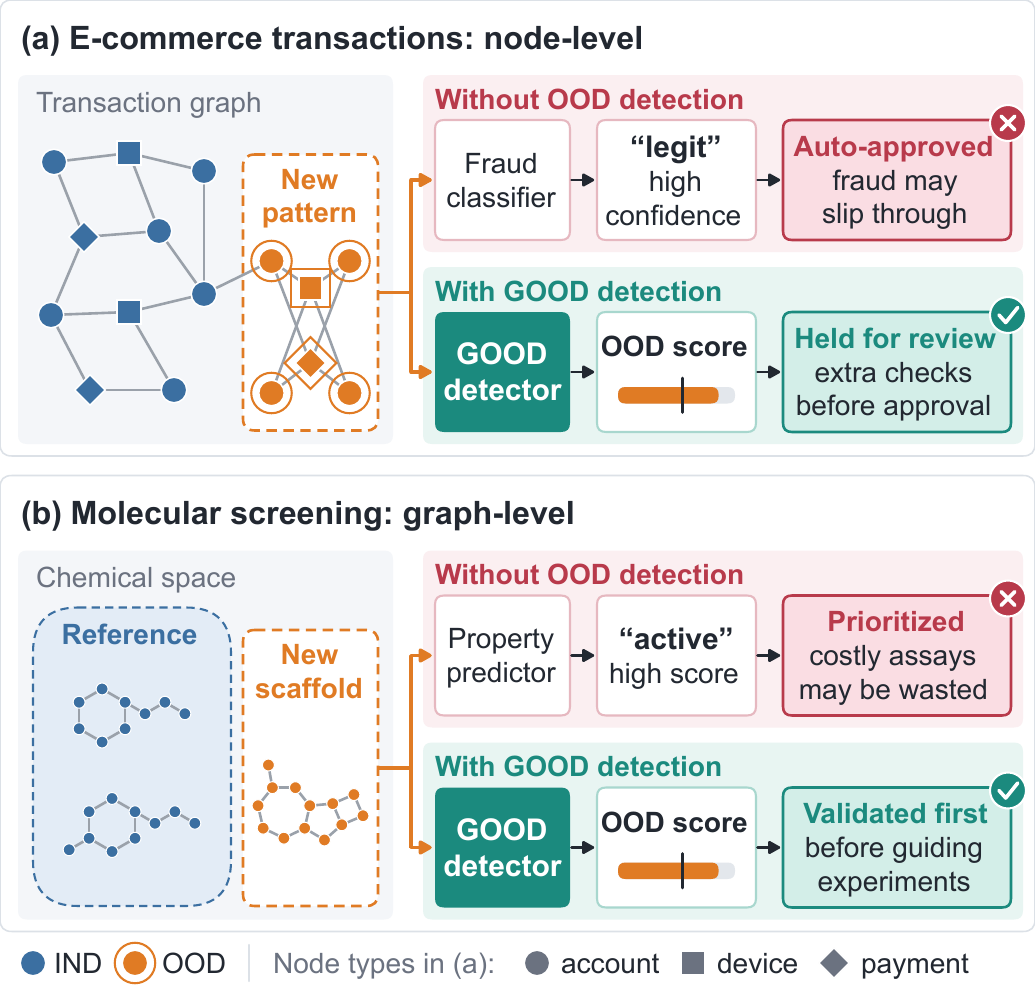}
    \caption{GOOD detection in open environments.
    (a) Node-level detection can flag unfamiliar transaction patterns for additional review.
    (b) Graph-level detection can flag molecular structures outside the reference chemical space for further validation.}
    \label{fig:good_motivation}
\end{figure}

To address this risk, \textbf{graph out-of-distribution (GOOD) detection} identifies nodes
or graphs outside the in-distribution (IND) population specified
for a task.
For flagged inputs, a system can defer an automatic decision and request further assessment.
Fig.~\ref{fig:good_motivation} illustrates this role at the node and
graph levels. In e-commerce, graph models assess transactions
through their links to accounts, devices, and payment
tokens~\cite{lu2022bright}.
When fraud patterns change, models trained on previously observed behavior may encounter transactions
that differ from their training data~\cite{zhang2022efraudcom}.
Node-level detection could flag such unfamiliar transactions for
additional checks before automated approval or rejection. At the
graph level, molecular screening poses a similar problem when
candidate compounds contain scaffolds absent from the training
collection~\cite{wu2018moleculenet,scalia2020evaluating,chen2025molecularuq}.
Graph-level detection could flag candidates outside the reference
chemical space for validation before their predicted properties
guide experimental priorities.

Despite extensive research on OOD detection, \textbf{graphs pose distinct
challenges because distribution shifts can affect both entities
and their relationships}~\cite{yang2024generalized,wu2023gnnsafe,GOOD-D}.
Such shifts may alter node attributes, connectivity, class
semantics, or dependencies between features and structure.
These changes can occur together: changing node behavior may
coincide with new interaction patterns, and unknown classes may
appear in domains with shifted features. This heterogeneity
complicates score design: scores based only on node attributes
can overlook relational changes, while classification confidence
may remain high under shifts that preserve class semantics.
Whether these changes should trigger rejection depends on the
detection target: source-based detection may treat class-preserving
changes as OOD, whereas semantic OOD detection under covariate
shift accepts shifted known-class samples~\cite{he2025gdda}.
GOOD detection requires sensitivity to the changes designated as
OOD while accepting the variations permitted by the task.

Even when the OOD target is fixed, \textbf{combining information
across a graph can strengthen OOD signals or obscure the
differences needed for detection}. At the node level,
neighborhood aggregation changes the representations used for
scoring, while score propagation combines detection scores after
they have been computed~\cite{wu2023gnnsafe,GRASP}. Because
connected nodes can have different IND/OOD memberships, these
operations may obscure an OOD node's novelty or make an IND node
appear unfamiliar. Class similarity provides limited guidance
for controlling this mixing: nodes from different known classes
can both be IND, and an OOD node can have predominantly IND
neighbors. At the graph level, pooling combines node or
substructure representations into a representation of the whole
graph. Novelty in a local substructure or in the arrangement of
familiar parts can be obscured if pooling emphasizes only patterns
useful for known-class prediction~\cite{GOOD-D,SGOOD}. In both
settings, reliable detection requires choosing which information
to combine and preserving the differences that support rejection.

Studies addressing these challenges have developed diverse methods
and reported improvements in node-level and graph-level OOD
detection~\cite{wu2023gnnsafe,GOOD-D}. However, common terminology
can obscure technical differences among these methods.
For example, \emph{energy-based approaches} propagate scores
derived from classification logits, fit explicit IND energy
models, or learn separation boundaries from generated
outliers~\cite{wu2023gnnsafe,anonymous2025DeGEM,wang2025GOLD}.
Comparing these approaches requires examining the learning
objectives and sources of OOD information behind their scores.
A separate difficulty arises when comparing reported performance:
studies differ in OOD targets, available labels, auxiliary data,
test-graph access, and threshold
selection~\cite{wu2023gnnsafe,GOOD-D,GRASP,xu2025glipood,lin2025cgod}.
Performance differences across studies can reflect both
methodological choices and evaluation conditions. A systematic
synthesis of these mechanisms and evaluation conditions is needed
to connect the research lines, interpret reported results, and
guide detector selection for specific tasks.

To provide this synthesis, we present a comprehensive and
systematic review of OOD detection on graphs.
We organize node-level and graph-level methods by their primary OOD
discrimination mechanisms and examine their information
requirements and evaluation conditions. We aim to establish a
common foundation for GOOD detection research and help shape
its future development. Our main contributions are as follows:
\begin{itemize}
    \item \textbf{A comprehensive review and mechanism-based taxonomy.}
    We organize 63 core methods into six families according to their
    primary OOD discrimination mechanisms: representation shaping,
    IND distribution modeling, predictive evidence modeling,
    relational evidence modeling, IND--OOD boundary learning, and
    detection calibration (Fig.~\ref{fig:taxonomy}). The review
    explains how these approaches construct OOD evidence and traces
    their technical development.

    \item \textbf{A comparison of method capabilities and evaluation settings.}
    We compare the supervision, reference data, test-time information,
    and calibration required by representative detectors. Our review
    of datasets, OOD constructions, benchmarks, and metrics examines
    how OOD targets and rejection criteria are defined, providing
    a basis for method selection and performance comparison.

    \item \textbf{Controlled comparisons of detection behavior.}
    We examine representative detectors on synthetic graphs under
    feature, structural, open-set semantic, and compound shifts
    across heterophilic, neutral, and homophilic settings.
    Comparisons with IND-based statistical scores identify
    distributional differences that some learned detector scores
    capture weakly. We assess IND classification, OOD ranking, and
    rejection at high IND retention as distinct performance
    criteria. The methodological review and these comparisons
    inform our discussion of open problems and future directions.
\end{itemize}

\noindent\textbf{Comparison to Existing Surveys.}
To the best of our knowledge, this is the first systematic survey dedicated to graph OOD detection.
Existing surveys examine general OOD detection~\cite{yang2024generalized,lu2025taskood}, graph OOD generalization and adaptation~\cite{li2025graphgeneralization,zhang2026graphshifts}, or the connections between classical OOD principles and GNN-based approaches~\cite{nguyen2026conceptualoodsurvey}. Our survey provides a comprehensive overview of node-level and graph-level OOD detection and a systematic taxonomy based on primary discrimination mechanisms. We discuss method capabilities, information requirements, and evaluation practices in depth, using controlled synthetic comparisons to interpret detector responses to different distribution shifts.
We identify open problems and research directions for advancing GOOD detection.

The remainder of this article is organized as follows. Section~\ref{sec:foundations} defines GOOD detection and introduces the taxonomy.
Sections~\ref{sec:R}--\ref{sec:C} review the six families of detection methods.
Section~\ref{sec:benchmarks} summarizes datasets, benchmarks, and evaluation practices.
Section~\ref{sec:good_comparison} compares method capabilities and analyzes detector behavior
under controlled distribution shifts.
Section~\ref{sec:future_directions} discusses future research directions and concludes the survey.

\section{Foundations of GOOD Detection}
\label{sec:foundations}


\subsection{Problem Definition}
\label{sec:Problem}

Let $\mathcal{G}=(\mathcal{V},\mathcal{E},\mathbf{A},\mathbf{X})$
denote an attributed graph with $n=|\mathcal{V}|$ nodes, where
$\mathcal{E}$ is the edge set, $\mathbf{A}\in\{0,1\}^{n\times n}$
is the adjacency matrix, and $\mathbf{X}\in\mathbb{R}^{n\times d}$
contains the $d$-dimensional node features. GOOD detection distinguishes IND and OOD samples
at the node or graph level.

\subsubsection{Detection Units}
In \emph{node-level detection}, the target is an individual node
within a graph. Its attributes and accessible graph relationships
provide information for detection. IND and OOD nodes can coexist
in the same graph and be connected to each
other~\cite{wu2023gnnsafe,GPN}. The detection label describes the
target node's membership, while its representation or score may
depend on neighbors with different IND/OOD memberships.

In \emph{graph-level detection}, each graph is evaluated as a complete
sample. Distributional differences can involve node attributes,
topology, or the composition and organization of
substructures~\cite{GOOD-D,SGOOD}. Familiar node types can form an
OOD graph through changes in their relationships. The detection
label refers to the whole graph, and node-level OOD labels may be
unavailable or undefined.

\subsubsection{IND Reference and OOD Targets}
Let $P_{\mathrm{in}}$ denote the reference distribution for the node
or graph population designated as IND. The application specifies
which departures from this reference constitute
OOD~\cite{yang2024generalized}. In open-set semantic detection,
the known classes form a label set $\mathcal{Y}_{\mathrm{in}}$,
and samples from classes outside this set are OOD. Other settings
target changes in node attributes, connectivity, or the dependence
between attributes and graph structure~\cite{wu2023gnnsafe}.
Changes across domains or time can combine these factors.


The same distributional change can receive different detection
labels, as shown in Table~\ref{tab:detection_target_contrast}.
A known-class graph from a new domain is IND in semantic OOD detection under
covariate shift~\cite{he2025gdda}; domain-based detection can
designate that new source as OOD.

\begin{table}[t]
\centering
\caption{Different detection targets for the same domain shift.}
\label{tab:detection_target_contrast}
\begingroup
\footnotesize
\setlength{\tabcolsep}{4pt}
\renewcommand{\arraystretch}{1.15}
\begin{tabularx}{\columnwidth}{@{}
    >{\raggedright\arraybackslash}p{.28\columnwidth}
    >{\raggedright\arraybackslash}X
    c@{}}
\toprule
\multicolumn{3}{@{}p{\columnwidth}@{}}{
    \emph{Input: a known-class graph from a new domain.}}\\
\midrule
Detection target & IND scope & Task label\\
\midrule
Semantic novelty
& Known classes; domain variation is accepted
& IND\\
Domain membership
& Graphs from the reference domain
& OOD\\
\bottomrule
\end{tabularx}
\endgroup
\end{table}

\begin{definition}[GOOD Detection]
Given an IND reference distribution $P_{\mathrm{in}}$ and a
specified OOD target, GOOD detection aims to distinguish IND test
samples from samples associated with an OOD distribution
$P_{\mathrm{out}}$.

Let $z_i$ denote a test node or graph, with $o_i=0$ for an IND sample
and $o_i=1$ for an OOD sample. A detector parameterized by $\theta$
assigns an OOD score $S_{\theta}(z_i)\in\mathbb{R}$, where larger
values indicate stronger OOD evidence. The binary decision is
\begin{equation}
    \hat{o}_i=\mathbb{I}\left[S_{\theta}(z_i)>\tau\right],
    \label{eq:good_detecetion}
\end{equation}
where $\tau$ is the detection threshold and $\hat{o}_i=1$ indicates
an OOD prediction. For node-level detection, the score can be
written as $S_{\theta}(v_i;\mathcal{G}_{\mathrm{obs}})$, where
$\mathcal{G}_{\mathrm{obs}}$ contains the graph information
available for inference. For graph-level detection, the scored
sample is the complete graph, $S_{\theta}(\mathcal{G}_i)$.
\end{definition}

\subsubsection{Information Available for Detection}
IND class labels are optional. Detectors can learn from unlabeled
reference graphs or use pretrained models and semantic
descriptions~\cite{GOOD-D,xu2025glipood}. Training data may include
samples whose IND/OOD membership is unknown, while auxiliary real
or synthetic outliers can provide explicit rejection
supervision~\cite{li2022GraphDE,HGOE2024mm}. In transductive
node detection, the attributes and edges of query nodes may be
available during learning even when their class labels and IND/OOD
memberships are unknown. Inductive detection evaluates nodes or
graphs unavailable during training.

Known-IND reference samples, held-out calibration data, or unlabeled
test samples can support score adjustment and threshold
selection~\cite{GRASP,lin2025cgod,hou2025d2go}. The threshold
determines the balance between rejecting IND samples and accepting
OOD samples. A detection setting specifies the data available for
fitting, model selection, calibration, and inference.

\subsubsection{Detection and Prediction}
GOOD detectors may operate independently or alongside an IND
classifier, which assigns class labels to samples accepted as IND.
Combining unknown-class rejection with known-class classification
gives the open-set recognition setting~\cite{yang2024generalized}.
The definition applies to GNNs, graph Transformers, and other
graph-based models. Graph OOD generalization evaluates predictive
performance under distribution shifts, while GOOD detection
identifies the inputs designated as OOD for the
task~\cite{wu2023gnnsafe,fang2024learnability}.\footnote{Graph backdoor defense targets
attack-induced behavior. DMGNN~\cite{sui2026dmgnn} uses
explanation-guided pruning to remove triggers; its IND/OOD
distinction concerns trigger features relative to clean data.}

\begin{remark}[Relation to graph anomaly detection]
Graph anomaly detection identifies departures from attribute or
relational patterns regarded as normal for an application. GOOD
detection labels samples by membership in a specified IND population,
including distinctions based on class semantics or data source.
The tasks share a detection target when the normal population defines
IND and the anomalies define OOD. UB-GOLD uses this connection to compare
graph-level anomaly and OOD detectors within a common
evaluation~\cite{wang2025ubgold}. A new domain can contain samples
regarded as normal for an application but designated OOD in a
source-based evaluation. Comparisons between the tasks should specify
the reference population, label construction, and available information.
\end{remark}


\begin{figure*}[t]
\centering
\begingroup
\definecolor{goodtaxroot}{HTML}{385B78}
\definecolor{goodtaxedge}{HTML}{8B939C}
\definecolor{goodtaxtext}{HTML}{263746}
\definecolor{goodtaxRfill}{HTML}{DCE5EF}
\definecolor{goodtaxRdraw}{HTML}{748AA6}
\definecolor{goodtaxRsub}{HTML}{EAF0F6}
\definecolor{goodtaxDfill}{HTML}{F2E9D9}
\definecolor{goodtaxDdraw}{HTML}{B89E68}
\definecolor{goodtaxDsub}{HTML}{F8F2E8}
\definecolor{goodtaxPfill}{HTML}{DFE8DD}
\definecolor{goodtaxPdraw}{HTML}{7F987A}
\definecolor{goodtaxPsub}{HTML}{EDF3EC}
\definecolor{goodtaxLfill}{HTML}{F0E2D8}
\definecolor{goodtaxLdraw}{HTML}{B68267}
\definecolor{goodtaxLsub}{HTML}{F8EEE8}
\definecolor{goodtaxBfill}{HTML}{ECDDE6}
\definecolor{goodtaxBdraw}{HTML}{A57990}
\definecolor{goodtaxBsub}{HTML}{F5ECF1}
\definecolor{goodtaxCfill}{HTML}{E4E0EF}
\definecolor{goodtaxCdraw}{HTML}{9183AA}
\definecolor{goodtaxCsub}{HTML}{F1EDF7}
\resizebox{\textwidth}{!}{%
\begin{forest}
for tree={
    grow=east,
    anchor=west,
    parent anchor=east,
    child anchor=west,
    edge={draw=goodtaxedge,line width=0.6pt},
    draw=goodtaxedge,
    rounded corners=3pt,
    align=left,
    l sep+=5pt,
    s sep=3pt,
    inner sep=3.5pt,
    outer sep=0pt,
    text=goodtaxtext,
    font=\footnotesize,
    minimum height=7mm
}
[\parbox{1.55cm}{\centering\bfseries GOOD\\Detection\\[2pt]},
    fill=goodtaxroot,draw=goodtaxroot,text=white,line width=0.8pt
  [\parbox{3.0cm}{\centering\bfseries Detection\\Calibration\\[2pt]\normalfont (3 studies)}, tier=family,fill=goodtaxCfill,draw=goodtaxCdraw,line width=0.8pt
    [\parbox{2.8cm}{\centering Dictionary-based Calibration}, tier=route,fill=goodtaxCsub,draw=goodtaxCdraw,line width=0.7pt
      [\parbox{7.7cm}{\raggedright D2GO~\cite{hou2025d2go}, BaCa~\cite{hou2026baca}}, tier=methods,fill=white,draw=goodtaxCdraw,line width=0.7pt]
    ]
    [\parbox{2.8cm}{\centering Conformal Calibration}, tier=route,fill=goodtaxCsub,draw=goodtaxCdraw,line width=0.7pt
      [\parbox{7.7cm}{\raggedright CGOD~\cite{lin2025cgod}}, tier=methods,fill=white,draw=goodtaxCdraw,line width=0.7pt]
    ]
  ]
  [\parbox{3.0cm}{\centering\bfseries IND--OOD\\Boundary Learning\\[2pt]\normalfont (15 studies)}, tier=family,fill=goodtaxBfill,draw=goodtaxBdraw,line width=0.8pt
    [\parbox{2.8cm}{\centering Meta-learning}, tier=route,fill=goodtaxBsub,draw=goodtaxBdraw,line width=0.7pt
      [\parbox{7.7cm}{\raggedright SMUG~\cite{SMUG}}, tier=methods,fill=white,draw=goodtaxBdraw,line width=0.7pt]
    ]
    [\parbox{2.8cm}{\centering Synthetic Outlier Generation}, tier=route,fill=goodtaxBsub,draw=goodtaxBdraw,line width=0.7pt
      [\parbox{7.7cm}{\raggedright GOLD~\cite{wang2025GOLD}, SynCo-OOD~\cite{wang2026synco}, NOAOM~\cite{zhou2026noaom}, PGR-MOOD~\cite{PGR-MOOD}, PGOS~\cite{sun2026pgos}, GDDA~\cite{he2025gdda}, SA-Diff~\cite{dong2026sadiff}, LLMGuard~\cite{lv2025llmguard}, LECT~\cite{ma2026lect}}, tier=methods,fill=white,draw=goodtaxBdraw,line width=0.7pt]
    ]
    [\parbox{2.8cm}{\centering Weakly Supervised Detection}, tier=route,fill=goodtaxBsub,draw=goodtaxBdraw,line width=0.7pt
      [\parbox{7.7cm}{\raggedright Open-WRF~\cite{Open-WRF}, Wang et al.~\cite{wang2026hyperspectralgood}, LLM-GOOD~\cite{xu2025llmgood}, CFC~\cite{ma2026cfc}}, tier=methods,fill=white,draw=goodtaxBdraw,line width=0.7pt]
    ]
    [\parbox{2.8cm}{\centering Outlier Exposure}, tier=route,fill=goodtaxBsub,draw=goodtaxBdraw,line width=0.7pt
      [\parbox{7.7cm}{\raggedright HGOE~\cite{HGOE2024mm}}, tier=methods,fill=white,draw=goodtaxBdraw,line width=0.7pt]
    ]
  ]
  [\parbox{3.0cm}{\centering\bfseries Relational Evidence\\Modeling\\[2pt]\normalfont (11 studies)}, tier=family,fill=goodtaxLfill,draw=goodtaxLdraw,line width=0.8pt
    [\parbox{2.8cm}{\centering Spectral Filtering}, tier=route,fill=goodtaxLsub,draw=goodtaxLdraw,line width=0.7pt
      [\parbox{7.7cm}{\raggedright SpecGap~\cite{gu2025spectralgap}}, tier=methods,fill=white,draw=goodtaxLdraw,line width=0.7pt]
    ]
    [\parbox{2.8cm}{\centering Neighborhood Discrepancy}, tier=route,fill=goodtaxLsub,draw=goodtaxLdraw,line width=0.7pt
      [\parbox{7.7cm}{\raggedright TopoOOD~\cite{TopoOOD}, RAGNOR~\cite{gu2025ragnor}}, tier=methods,fill=white,draw=goodtaxLdraw,line width=0.7pt]
    ]
    [\parbox{2.8cm}{\centering Propagation-based Detection}, tier=route,fill=goodtaxLsub,draw=goodtaxLdraw,line width=0.7pt
      [\parbox{7.7cm}{\raggedright GNNSafe~\cite{wu2023gnnsafe}, GRASP~\cite{GRASP}, EDBD~\cite{anonymous2025EDBD}, EPGNN~\cite{zhang2025epgnn}, OODHG~\cite{yin2025oodhg}, GNSD~\cite{GNSD}}, tier=methods,fill=white,draw=goodtaxLdraw,line width=0.7pt]
    ]
    [\parbox{2.8cm}{\centering OOD-aware Message Passing}, tier=route,fill=goodtaxLsub,draw=goodtaxLdraw,line width=0.7pt
      [\parbox{7.7cm}{\raggedright LMN~\cite{LMN}, OODGAT~\cite{song2022OODGAT}}, tier=methods,fill=white,draw=goodtaxLdraw,line width=0.7pt]
    ]
  ]
  [\parbox{3.0cm}{\centering\bfseries Predictive Evidence\\Modeling\\[2pt]\normalfont (8 studies)}, tier=family,fill=goodtaxPfill,draw=goodtaxPdraw,line width=0.8pt
    [\parbox{2.8cm}{\centering Semantic Matching}, tier=route,fill=goodtaxPsub,draw=goodtaxPdraw,line width=0.7pt
      [\parbox{7.7cm}{\raggedright GLIP-OOD~\cite{xu2025glipood}}, tier=methods,fill=white,draw=goodtaxPdraw,line width=0.7pt]
    ]
    [\parbox{2.8cm}{\centering Energy Scores}, tier=route,fill=goodtaxPsub,draw=goodtaxPdraw,line width=0.7pt
      [\parbox{7.7cm}{\raggedright NODESAFE~\cite{NODESAFE}, ML-GOOD~\cite{cai2025mlgood}, GeoEnergy~\cite{wang2025geoenergy}}, tier=methods,fill=white,draw=goodtaxPdraw,line width=0.7pt]
    ]
    [\parbox{2.8cm}{\centering Uncertainty Estimation}, tier=route,fill=goodtaxPsub,draw=goodtaxPdraw,line width=0.7pt
      [\parbox{7.7cm}{\raggedright GKDE~\cite{GKDE}, S-BGCN-T-K~\cite{S-BGCN-T-K}, OpenWGL~\cite{OpenWGL}, ML-EGNNs~\cite{ML-EGNNs}}, tier=methods,fill=white,draw=goodtaxPdraw,line width=0.7pt]
    ]
  ]
  [\parbox{3.0cm}{\centering\bfseries IND Distribution\\Modeling\\[2pt]\normalfont (4 studies)}, tier=family,fill=goodtaxDfill,draw=goodtaxDdraw,line width=0.8pt
    [\parbox{7.7cm}{\raggedright GraphDE~\cite{li2022GraphDE}, GPN~\cite{GPN}, EnergyDef~\cite{EnergyDef2024kdd}, DeGEM~\cite{anonymous2025DeGEM}}, tier=methods,fill=white,draw=goodtaxDdraw,line width=0.7pt]
  ]
  [\parbox{3.0cm}{\centering\bfseries Representation\\Shaping\\[2pt]\normalfont (22 studies)}, tier=family,fill=goodtaxRfill,draw=goodtaxRdraw,line width=0.8pt
    [\parbox{2.8cm}{\centering Temporal Representation Learning}, tier=route,fill=goodtaxRsub,draw=goodtaxRdraw,line width=0.7pt
      [\parbox{7.7cm}{\raggedright OSSC~\cite{OSSC}}, tier=methods,fill=white,draw=goodtaxRdraw,line width=0.7pt]
    ]
    [\parbox{2.8cm}{\centering Embedding Geometry}, tier=route,fill=goodtaxRsub,draw=goodtaxRdraw,line width=0.7pt
      [\parbox{7.7cm}{\raggedright UGNN~\cite{UGNN}, GPN-CE-GD~\cite{GPN-CE-GD}, HGOOD-D~\cite{ding2026hgoodd}, UniGOD~\cite{song2026unigod}, Feature norm regularization~\cite{zhang2026featurenormood}}, tier=methods,fill=white,draw=goodtaxRdraw,line width=0.7pt]
    ]
    [\parbox{2.8cm}{\centering Graph Prompting}, tier=route,fill=goodtaxRsub,draw=goodtaxRdraw,line width=0.7pt
      [\parbox{7.7cm}{\raggedright AAGOD~\cite{guo2023AAGOD}, DGP~\cite{yang2026dgp}, IBPL~\cite{cao2025ibpl}, SIGOOD~\cite{wang2026sigood}, APEKG~\cite{li2026apekg}}, tier=methods,fill=white,draw=goodtaxRdraw,line width=0.7pt]
    ]
    [\parbox{2.8cm}{\centering Information Bottleneck}, tier=route,fill=goodtaxRsub,draw=goodtaxRdraw,line width=0.7pt
      [\parbox{7.7cm}{\raggedright GOODAT~\cite{wang2024GOODAT}, RedOUT~\cite{hou2025redout}, TIDE~\cite{wang2026tide}}, tier=methods,fill=white,draw=goodtaxRdraw,line width=0.7pt]
    ]
    [\parbox{2.8cm}{\centering Multi-view Representation Learning}, tier=route,fill=goodtaxRsub,draw=goodtaxRdraw,line width=0.7pt
      [\parbox{7.7cm}{\raggedright GOOD-D~\cite{GOOD-D}, SEGO~\cite{hou2025sego}, SGOOD~\cite{SGOOD}, KEGOD~\cite{wang2026kegod}, CLINIC~\cite{wang2026clinic}, HGOOD~\cite{fan2026hgood}, UniGOOD~\cite{liao2026unigood}, TNT-OOD~\cite{wang2025tntood}}, tier=methods,fill=white,draw=goodtaxRdraw,line width=0.7pt]
    ]
  ]
]
\end{forest}%
}
\endgroup
\caption{Taxonomy of graph out-of-distribution (GOOD) detection methods. Repository: \protect\url{https://github.com/ca1man-2022/Awesome-GOOD-Detection}.}
\label{fig:taxonomy}
\end{figure*}

\subsection{Taxonomy of GOOD Detection Methods}
\label{sec:taxonomy}

We organize the 63 surveyed methods into six families according to
their primary OOD discrimination mechanisms, as shown in Fig.~\ref{fig:taxonomy}.
Different OOD targets require sensitivity to different changes in node attributes, connectivity, and class
semantics. Learning objectives and inference operations specify how
information about these changes enters detection. Graph relations
can shape representations or act on evidence that has already been
computed. Organizing methods by their primary mechanism links these
design choices across architectures and detection units.

Shared model components can serve different detection purposes.
DeGEM fits an explicit energy model to IND embeddings, NODESAFE
constrains classification logits to improve the resulting energy
scores, and GOLD learns separation from generated
outliers~\cite{anonymous2025DeGEM,NODESAFE,wang2025GOLD}.
Their differences lie in what is learned and how test inputs are
assessed, even though all three methods use energy scores.
For methods that combine several mechanisms, the assignment follows
their principal contribution to OOD discrimination, as expressed in
the proposed learning and inference operations. GPN's class-conditional
latent densities supply the evidence used in graph propagation;
GNNSafe propagates energies derived from a
classifier~\cite{GPN,wu2023gnnsafe}. We classify their main
contributions as IND Distribution Modeling and Relational Evidence
Modeling, respectively. Supervision,
detection granularity, and reference-data access are compared
separately as conditions under which each method operates.

\noindent\textbf{Representation Shaping.}
Learning objectives and input adaptations shape node and graph embeddings for OOD detection. Multi-view representation learning combines complementary graph descriptions; information bottlenecks select what the encoder retains; graph prompting adapts the input to a pretrained model. Geometric constraints organize embeddings through prototypes, distances, curvature, or feature norms, while temporal models encode changes across graph snapshots~\cite{GOOD-D,wang2026tide,guo2023AAGOD,GPN-CE-GD,OSSC}.

\noindent\textbf{IND Distribution Modeling.}
These methods model the IND distribution in graph data or learned latent representations.
Source mixture models infer whether a test sample belongs to the IND component, density models measure support in the embedding space, and energy-based models estimate unnormalized distributions~\cite{li2022GraphDE,GPN,anonymous2025DeGEM}. Detection uses the resulting source posteriors, density-derived uncertainty, or learned energy values.

\noindent\textbf{Predictive Evidence Modeling.}
Predictive uncertainty, classification logits, and graph--text similarity provide detection scores. Uncertainty estimation quantifies support for known labels, energy scores summarize prediction outputs, and semantic matching compares graph representations with class descriptions~\cite{GKDE,NODESAFE,xu2025glipood}.

\noindent\textbf{Relational Evidence Modeling.}
Detection relies on neighborhood discrepancies, OOD-specific propagation, or topological differences.
OOD-aware message passing regulates neighboring contributions; propagation combines scores or uncertainty across connected nodes. Neighborhood discrepancy measures local disagreement, while spectral filtering uses graph-wide structure to modify the features used for detection~\cite{song2022OODGAT,wu2023gnnsafe,TopoOOD,gu2025spectralgap}.

\noindent\textbf{IND--OOD Boundary Learning.}
Observed or synthesized outliers and surrogate IND/OOD labels guide the learning of rejection boundaries. Outlier exposure, weakly supervised detection, and synthetic outlier generation supply examples or labels for discriminative learning. Meta-learning treats selected observed classes as unknown within training tasks~\cite{HGOE2024mm,Open-WRF,wang2025GOLD,SMUG}.

\noindent\textbf{Detection Calibration.}
Reference data support score adjustment and threshold selection. Conformal calibration compares test scores with held-out IND scores and controls marginal IND false rejection under its statistical assumptions~\cite{lin2025cgod}. Dictionary-based calibration uses evolving IND/OOD reference sets to refine scores at test time~\cite{hou2025d2go,hou2026baca}.



\section{Representation Shaping}
\label{sec:R}

\begin{figure}[t]
    \centering
    \includegraphics[width=\columnwidth]{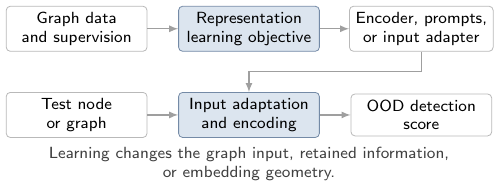}
    \caption{Representation shaping for GOOD detection.}
    \label{fig:representation_framework}
    \label{fig:R method}
\end{figure}

Representation shaping learns or adapts node and graph embeddings to make OOD inputs distinguishable from IND samples (Fig.~\ref{fig:representation_framework}). Methods differ in the graph information retained by the encoder and the organization of the embedding space.

\subsection{Multi-view Representation Learning}
\label{sec:representation_multiview}
\label{sec:representation_contrastive}
\label{sec:representation_substructure}
\label{sec:representation_semantic}

Multi-view methods learn IND patterns from complementary descriptions of graph data, including structural scales, encoding branches, and information modalities. \textbf{GOOD-D}~\cite{GOOD-D} uses perturbation-free views with contrastive objectives at the node, graph, and group levels. Inconsistency across these levels supports graph-level OOD detection without IND class labels. \textbf{SEGO}~\cite{hou2025sego} introduces a coding-tree view obtained by minimizing structural entropy and aligns local, global, and tree-level representations.

Substructure-based views preserve local patterns and their organization. \textbf{SGOOD}~\cite{SGOOD} constructs a supergraph of substructures and jointly encodes it with the original graph, using substructure-preserving augmentation to retain local semantics. \textbf{KEGOD}~\cite{wang2026kegod} combines neural and graph-kernel branches to capture implicit and explicit substructure information. It learns IND representations through cross-branch consistency and detects OOD graphs using discrepancies between branches. These approaches differ in how they describe substructures: explicit relationships among them in SGOOD, and complementary encoders in KEGOD.

Other views capture relationships beyond local substructures. \textbf{CLINIC}~\cite{wang2026clinic} constructs an affinity graph among graph samples and uses a meta-graph neural network to model higher-order affinities. Its learning objective aligns affinity distributions across views and the representations of contextually related samples. It also compresses these representations to improve separation between IND and OOD graphs. \textbf{HGOOD}~\cite{fan2026hgood} combines a graph-feature branch with a hypergraph-based global branch, aligning them through prototype contrast to represent ordinary graphs. \textbf{UniGOOD}~\cite{liao2026unigood} learns from invariant subgraphs through distributional subgraph generation, spectral contrastive learning, and invariance regularization.

For text-rich networks, the complementary views come from node text and topology. \textbf{TNT-OOD}~\cite{wang2025tntood} uses cross-attention to incorporate topology into textual representations and a hypernetwork to generate node-specific transformations. The resulting alignment adapts the interaction between textual content and relational context for node-level detection.

The choice of views specifies the discrepancies emphasized during
learning. Multi-scale agreement relates local and global graph
descriptions; cross-encoder agreement compares representations of
the same graph; text--topology alignment relates node semantics to
its context. Each choice defines a different notion of consistency
within IND data, even when the methods use similar alignment
objectives.

\subsection{Information Bottleneck}
\label{sec:representation_bottleneck}

Information bottlenecks seek compact representations that retain information specified by the learning objective. \textbf{GOODAT}~\cite{wang2024GOODAT} learns a test-time graph masker to extract informative subgraphs while keeping the pretrained GNN fixed and requiring no access to its original training data. \textbf{RedOUT}~\cite{hou2025redout} targets structural redundancy through a redundancy-aware graph information bottleneck and structural-entropy minimization at test time.

At the node level, \textbf{TIDE}~\cite{wang2026tide} decomposes representations into feature-specific, structure-specific, and joint information~\cite{mondal2025variational}. Its bottleneck retains label-relevant joint information while suppressing potentially spurious individual components. The selection concerns the contributions of attributes and connectivity, extending beyond the choice of an informative subgraph.

\subsection{Graph Prompting}
\label{sec:representation_prompt}

Graph prompts modify the input to a pretrained encoder to improve OOD detection. \textbf{AAGOD}~\cite{guo2023AAGOD} learns graph-specific adjacency amplifiers using IND data. \textbf{DGP}~\cite{yang2026dgp} adjusts edge weights through class-specific and class-agnostic prompts, capturing complementary IND patterns while reusing a fixed encoder. Both change how graph connectivity contributes to the resulting embeddings.

Prompt optimization can draw on different training signals. \textbf{IBPL}~\cite{cao2025ibpl} prompts node attributes and graph structure jointly; its information bottleneck reduces overlap with perturbed graphs while preserving task-relevant content. \textbf{SIGOOD}~\cite{wang2026sigood} iteratively refines test-time prompts using energy preferences between original and prompt-enhanced graphs. For molecular graphs, \textbf{APEKG}~\cite{li2026apekg} combines attribute- and structure-aware prompts with knowledge from a pretrained molecular model, using node- and graph-level contrastive learning to align representations. These choices determine what supervision or pretrained knowledge guides the prompt and whether adaptation continues at test time.

\subsection{Embedding Geometry}
\label{sec:representation_geometry}

Geometric approaches organize embeddings through prototypes, distances, curvature, or feature magnitude. \textbf{UGNN}~\cite{UGNN} learns from labeled known-class graphs and an unlabeled pool that may include unknown classes. Multiple subgraph views estimate graph certainty, while prototypes and posterior assignments organize known and unknown categories. \textbf{GPN-CE-GD}~\cite{GPN-CE-GD} adds distance-based regularization to GPN's uncertainty cross-entropy objective, constraining neighboring latent representations or evidence vectors. The motivation is that fitting known-class predictions alone can discard features that distinguish unknown classes. Its regularizer can use unlabeled nodes already present in the graph, including held-out classes. This encourages clustered OOD nodes to retain their organization without requiring their OOD labels.

\textbf{HGOOD-D}~\cite{ding2026hgoodd} combines an information-bottleneck graph extractor with hierarchical contrastive learning in hyperbolic space to represent semantic hierarchies. \textbf{UniGOD}~\cite{song2026unigod} learns manifold curvature and models node-embedding evolution through neural stochastic differential equations. \textbf{Feature norm regularization}~\cite{zhang2026featurenormood} controls embedding magnitude, limiting excessively large feature norms during training to stabilize representations.

\subsection{Temporal Representation Learning}
\label{sec:representation_temporal}

Temporal models use a node's history to represent changes in its attributes and structural context. \textbf{OSSC}~\cite{OSSC} combines graph convolution with a dynamic variational autoencoder, using stochastic and deterministic latent states to capture dependencies across snapshots. Node representations sampled from these states are trained with an open-world classification objective. At inference, OpenMax recognizes unknown-class nodes, while Softmax classifies known-class nodes.

\noindent\textbf{Discussion.}
The learning objective determines which changes remain visible after encoding. View alignment and information bottlenecks suppress some variation, while prompts and geometric constraints alter how samples are separated. Their suitability depends on the intended OOD target: invariance is useful for variations that should be accepted, whereas detection requires retaining the differences that should trigger rejection.

\section{IND Distribution Modeling}
\label{sec:D}
\label{sec:distribution_probabilistic}
\label{sec:distribution_latent_density}
\label{sec:distribution_energy}

\begin{figure}[t]
\centering
\includegraphics[width=\columnwidth]{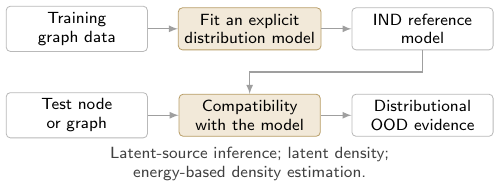}
\caption{IND distribution modeling for GOOD detection.}
\label{fig:distribution_framework}
\label{fig:D method}
\end{figure}

These methods detect OOD nodes or graphs by modeling the distribution of graph data or their embeddings (Fig.~\ref{fig:distribution_framework}). Detection uses a source posterior, a latent density, or an energy function fitted to the data.

A graph's source can be inferred as a latent variable. \textbf{GraphDE}~\cite{li2022GraphDE} introduces a binary environment variable into a generative model with IND and outlier components. During training, environment inference uses adjacency, node features, and class labels to identify and down-weight suspected outliers. At test time, the detector computes the environment posterior from adjacency and node features through the learned structure model.

\textbf{GPN}~\cite{GPN} obtains node-level evidence from class-conditional densities. Normalizing flows model these densities in a learned embedding space; the densities become Dirichlet pseudo-counts, which are propagated through personalized PageRank to produce posteriors incorporating neighboring evidence. Comparing uncertainty before and after propagation shows how network context changes feature-based support. The encoder and density heads are learned jointly through a Bayesian predictive objective. GraphDE directly infers distributional membership, while GPN converts latent densities into predictive uncertainty.

Energy-based models fit an unnormalized density, assigning lower energy to samples supported by the model. \textbf{EnergyDef}~\cite{EnergyDef2024kdd} uses an energy-based GNN for category-free node OOD detection. Contrastive divergence fits the energy function, while Langevin dynamics and graph structure estimation provide synthetic nodes for model fitting. The learned energy supplies the detection score.

\textbf{DeGEM}~\cite{anonymous2025DeGEM} separates topology-aware encoding from energy estimation for node OOD detection, including on heterophilic graphs. A multi-hop encoder produces continuous embeddings, and a separate energy head is trained through maximum likelihood with Markov chain Monte Carlo sampling. Conditional energy modeling and energy-guided readout are coordinated through recurrent updates. EnergyDef generates nodes together with estimated graph structure; DeGEM samples embeddings, avoiding discrete adjacency sampling while incorporating graph information through the encoder.

\noindent\textbf{Discussion.}
GraphDE infers source membership by comparing the IND and outlier
components of its structure model. GPN estimates density support
from feature-only embeddings before incorporating neighboring
evidence, while DeGEM evaluates embeddings that already contain
topology information~\cite{li2022GraphDE,GPN,anonymous2025DeGEM}.
This ordering determines whether graph context changes the input
to density estimation or the density-derived evidence used for
prediction.

Likelihood-based scores can be influenced by background statistics;
likelihood-ratio methods contrast a data model with a background model to reduce this influence~\cite{ren2019likelihood}.
The representation and modeled distribution jointly determine which departures receive low support.
During maximum-likelihood energy training, generated negatives estimate expectations under the model distribution;
boundary-learning outliers carry an explicit OOD rejection target~\cite{anonymous2025DeGEM,HGOE2024mm}.

\section{Predictive Evidence Modeling}
\label{sec:P}

\begin{figure}[t]
\centering
\includegraphics[width=\columnwidth]{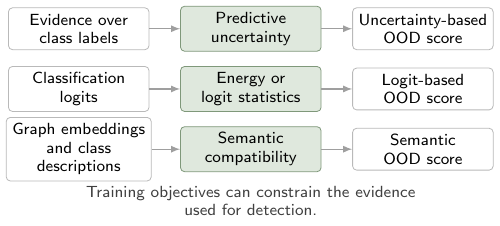}
\caption{Predictive evidence for GOOD detection.}
\label{fig:predictive_framework}
\label{fig:P method}
\end{figure}

These methods detect OOD samples using uncertainty about known labels, energies computed from classification logits, or similarity to semantic descriptions (Fig.~\ref{fig:predictive_framework}). They differ in how they measure insufficient support for a prediction.

\subsection{Uncertainty Estimation}
\label{sec:predictive_uncertainty}

Missing class evidence and conflicting evidence have different meanings for detection. \textbf{GKDE}~\cite{GKDE} collects labeled-node evidence through graph kernels and constructs Dirichlet opinions. Its subjective-logic formulation distinguishes \emph{vacuity}, representing insufficient evidence and supporting OOD detection, from \emph{dissonance}, representing disagreement among classes and helping identify misclassifications. For multimodal remote sensing, \textbf{S-BGCN-T-K}~\cite{S-BGCN-T-K} estimates uncertainty on a graph of pixels using Dirichlet opinions, teacher knowledge distillation, a graph-based prior, and dropout inference.

\textbf{OpenWGL}~\cite{OpenWGL} estimates predictive uncertainty through stochastic node representations. It trains a constrained variational graph autoencoder with label and class-uncertainty objectives. Predictions from multiple samples of each node's representation support unknown-class rejection through an automatically determined confidence threshold. Reconstruction helps train the encoder; predictive confidence determines rejection.

For labels that can occur simultaneously, \textbf{ML-EGNNs}~\cite{ML-EGNNs} predicts positive and negative evidence for each label, represents label-wise opinions with Beta distributions, and combines label beliefs into a joint belief for detection. Kernel-based positive-evidence estimation further refines node evidence. This preserves label-wise support when estimating uncertainty about a multi-label assignment.

\subsection{Energy Scores}
\label{sec:predictive_energy}

Classification logits provide OOD scores beyond normalized class probabilities. A common choice is the negative log-sum-exp energy~\cite{liu2020energy}:
\begin{equation}
    E_{\theta}(z)=-T\log\sum_{c=1}^{C}
    \exp\!\left(\frac{\ell_{\theta,c}(z)}{T}\right),
    \label{eq:predictive_energy}
\end{equation}
where $z$ is the evaluated node or graph, $\ell_{\theta,c}(z)$ is its logit for IND class $c$, and $T>0$ is the temperature. Under this sign convention, higher energy indicates stronger OOD evidence. This score is computed from prediction outputs; the models in Section~\ref{sec:D} fit explicit distributional energies.

Adding the same constant to a node's class logits changes its energy while preserving its class probabilities. \textbf{NODESAFE}~\cite{NODESAFE} addresses this sensitivity by constraining the variation of logit norms and logit sums across nodes during training. These constraints curb extreme negative-energy values and logit shifts before the scores undergo GNNSafe-style propagation.

\textbf{ML-GOOD}~\cite{cai2025mlgood} adapts energy scoring to multi-label node detection, combining the average and maximum of label-wise binary energies. \textbf{GeoEnergy}~\cite{wang2025geoenergy} changes the geometry of the logits: it normalizes class-weight vectors while preserving node-embedding norms, separating directional and magnitude contributions before computing energy. This transformation requires no retraining of the graph model.

\subsection{Semantic Matching}
\label{sec:predictive_semantic}

Textual descriptions provide semantic references for detection. \textbf{GLIP-OOD}~\cite{xu2025glipood} uses GraphCLIP~\cite{zhu2025graphclip}, whose GraphGPS encoder maps node-centered subgraphs into a shared graph--text representation space. Similarities to IND and OOD class descriptions yield zero-shot node OOD scores. When OOD class names are unavailable, an LLM proposes pseudo-OOD names. These names serve as semantic references during inference with the pretrained model, which requires no training on task-specific labeled nodes. The scores depend on graph--text alignment and the choice of candidate descriptions.

\noindent\textbf{Discussion.}
These methods measure support relative to a label space or a set of semantic descriptions. Their sensitivity to feature and structural shifts depends on how those changes affect uncertainty, logits, or graph--text similarity. This sensitivity should be evaluated separately from known-class prediction accuracy.

\section{Relational Evidence Modeling}
\label{sec:S}
\label{sec:L}

\begin{figure}[t]
\centering
\includegraphics[width=\columnwidth]{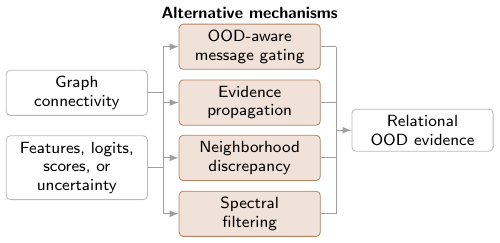}
\caption{Relational evidence for GOOD detection.}
\label{fig:relational_framework}
\label{fig:S method}
\end{figure}

Relational methods use graph connections to determine how information is exchanged or compared for OOD detection (Fig.~\ref{fig:relational_framework}). They regulate messages between IND and OOD nodes, propagate scores or uncertainty, and use neighborhood or spectral properties to construct detection scores.

\subsection{OOD-aware Message Passing}
\label{sec:structure_gating}

Connections between IND and OOD nodes can mix conflicting information during message passing. \textbf{LMN}~\cite{LMN} introduces latent IND/OOD indicators within a variational formulation and learns neighborhood mixing weights through bilevel optimization. These weights attenuate messages from suspected OOD neighbors. \textbf{OODGAT}~\cite{song2022OODGAT} incorporates OOD awareness into graph attention, jointly regulating information exchange, detecting unknown nodes, and predicting known classes. Both methods use inferred IND/OOD membership to decide how neighboring attributes contribute to a node's representation.

\subsection{Propagation-based Detection}
\label{sec:structure_propagation}

Propagation combines detection information across connected nodes. A representative update for scalar OOD scores is
\begin{equation}
\label{eq:structural_score_refinement}
    s_i^{(t+1)}=(1-\alpha_i)s_i^{(t)}+
    \alpha_i\sum_{j\in\mathcal{N}(i)}w_{ij}s_j^{(t)},
\end{equation}
where larger scores indicate stronger OOD evidence, $\alpha_i\in[0,1]$ controls neighborhood influence, and $w_{ij}\geq0$ with $\sum_{j\in\mathcal{N}(i)}w_{ij}=1$.

\textbf{GNNSafe}~\cite{wu2023gnnsafe} starts from energies computed from classification logits and propagates them without additional training. Its base detector uses IND supervision; an outlier-exposure variant adds energy regularization. \textbf{GRASP}~\cite{GRASP} relates score separation to connections within and across IND/OOD sources, then augments the propagation graph using selected known-IND training nodes. NODESAFE's constraints on logits address the initial scores supplied to propagation (Section~\ref{sec:P})~\cite{NODESAFE}.

Selective propagation limits mixing across estimated IND/OOD boundaries. For OOD states spreading over a fixed graph, \textbf{EDBD}~\cite{anonymous2025EDBD} uses energy similarity to weight transfer across edges and neighborhood consistency to adjust aggregation strength. \textbf{EPGNN}~\cite{zhang2025epgnn} combines an energy evaluator with propagation that reduces transfer between nodes with dissimilar energies; training aligns the energy evaluations with predictive entropy. \textbf{OODHG}~\cite{yin2025oodhg} extends energy propagation to heterogeneous graphs through meta-paths and an energy constraint, allowing node and edge types to determine which neighbors contribute.

\textbf{GNSD}~\cite{GNSD} models evolving node states through stochastic graph diffusion. A drift network describes predictive dynamics, while a stochastic forcing network models the propagation of epistemic uncertainty. Detection uses uncertainty estimated from these trajectories. Here propagation operates within uncertainty estimation.

\subsection{Neighborhood Discrepancy}
\label{sec:structure_topology}
\label{sec:structure_discrepancy}

Local disagreement provides another source of OOD information. \textbf{TopoOOD}~\cite{TopoOOD} computes generalized node-wise Dirichlet energy from GNN logits and edges within a node's $k$-hop neighborhood. Structural-proximity weights preserve the central node's context as neighborhoods expand. The score measures variation of the logit signal over connected nodes. Changes in node predictions can alter this score even when the graph remains fixed. \textbf{RAGNOR}~\cite{gu2025ragnor} instead uses embedding norms, applying global normalization and comparing each node with median-based neighborhood references. Multi-hop blending extends the local context. It supports node- and graph-level detection without retraining the backbone. TopoOOD measures pairwise variation within neighborhoods; RAGNOR measures deviation from a robust local reference.

\subsection{Spectral Filtering}
\label{sec:structure_spectral}

\textbf{SpecGap}~\cite{gu2025spectralgap} uses graph-wide spectral properties to modify high-level features for graph-level detection. For the combinatorial Laplacian $\mathbf{L}=\mathbf{D}-\mathbf{A}$ with $0=\lambda_1\leq\cdots\leq\lambda_n$, it defines the gap as $\lambda_n-\lambda_{n-1}$. The method subtracts the feature component along the eigenvector associated with $\lambda_{n-1}$, scaled by this gap. Subsequent model layers use these adjusted features, introducing a parameter-free correction based on the input graph's spectrum.

A related extension, \textbf{HyperGOOD}~\cite{cai2026hypergood} combines multi-scale spectral decomposition, energy scoring, and uncertainty propagation to detect OOD nodes or hyperedges in native hypergraphs.
Its input is a hypergraph, and HGOOD (Section~\ref{sec:R}) constructs hypergraph views for ordinary-graph detection~\cite{fan2026hgood}.

\noindent\textbf{Discussion.}
Relational processing changes detection at different points.
OOD-aware message passing changes node representations before scoring;
errors in inferred IND/OOD membership can affect both the representation
and the resulting score. Post-hoc score propagation keeps the trained
representation fixed and combines an existing detection signal across
neighbors. Its benefit depends on initial score separation and the
balance of connections within and across IND/OOD sources~\cite{GRASP}.
Class homophily describes label similarity, while IND/OOD connectivity
describes how detection sources mix.

Neighborhood discrepancy uses local variation as a detection signal.
The choice of signal and reference determines whether this variation
reflects the intended OOD change or familiar differences among IND
nodes~\cite{TopoOOD,gu2025ragnor}. Spectral filtering acts on the signal
itself, using graph-wide structure to change the features passed to
later model layers~\cite{gu2025spectralgap}.

\section{IND--OOD Boundary Learning}
\label{sec:O}
\label{sec:B}

\begin{figure}[t]
\centering
\includegraphics[width=\columnwidth]{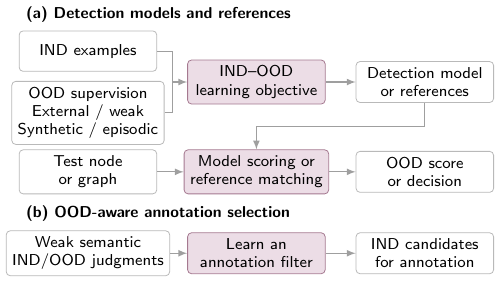}
\caption{IND--OOD boundary learning with outlier guidance.}
\label{fig:boundary_framework}
\label{fig:O method}
\end{figure}

This family learns to separate IND data from observed or constructed outliers (Fig.~\ref{fig:boundary_framework}). OOD supervision may train the detector, a filter for selecting IND annotations, or reference prototypes. Its source determines which departures receive explicit rejection supervision.

\subsection{Outlier Exposure}
\label{sec:outlier_exposure}

Outlier exposure trains detectors against auxiliary outliers to
learn rejection cues that can transfer to unseen OOD
inputs~\cite{hendrycks2019outlierexposure,bitterwolf2022breaking}.
\textbf{HGOE}~\cite{HGOE2024mm} combines graphs from other datasets with internal pseudo-outliers generated from IND subgroups. It estimates subgroup graphons and samples intermediate graph structures, combining external diversity with negatives derived from the IND collection. A boundary-aware exposure loss weights individual outliers according to their informativeness, concentrating learning on useful IND--OOD distinctions.

\subsection{Weakly Supervised Detection}
\label{sec:outlier_weak}

Weak IND/OOD labels can be obtained from an initial detector or semantic judgments. \textbf{Open-WRF}~\cite{Open-WRF} ranks nodes using existing OOD scores and assigns weak labels according to an expected OOD proportion. A two-layer GCN then learns a binary rejection rule. The paper also introduces GOOD for neighborhood score aggregation. In hyperspectral imagery, Wang et al.~\cite{wang2026hyperspectralgood} use a collaborative-representation detector to select candidate background and anomalous pixels within one scene. The resulting pseudo IND/OOD labels train a graph detector with separate energy margins; propagation over a $k$-nearest-neighbor graph refines its scores. The detection target is a departure from the scene's background.

For text-attributed graphs, \textbf{LLM-GOOD}~\cite{xu2025llmgood} obtains noisy IND/OOD judgments from an LLM on a small subset of nodes. A lightweight GNN filter learns from these judgments and selects informative likely-IND nodes for human annotation. Verified labels train the target classifier. \textbf{CFC}~\cite{ma2026cfc} uses an LLM to identify observed unknown-class candidates and propose labels, followed by a fine-grained GNN classifier and further refinement of unknown-class predictions. LLM-GOOD uses weak labels to allocate IND annotation effort, while CFC uses observed OOD candidates in subsequent recognition.

\subsection{Synthetic Outlier Generation}
\label{sec:outlier_synthesis}

Synthetic negatives can be constructed in latent space or as explicit nodes and graphs. \textbf{GOLD}~\cite{wang2025GOLD} alternates between a generator that tracks evolving IND node embeddings and a detector that increases their energy separation from generated embeddings. \textbf{SynCo-OOD}~\cite{wang2026synco} attracts IND node embeddings toward a shared center and repels synthetic negatives through contrastive learning, then detects OOD nodes using cosine similarity to that center. \textbf{NOAOM}~\cite{zhou2026noaom} targets near-OOD graphs by iteratively refining latent graph representations around IND prototypes, balancing proximity, separation, and diversity.
Latent synthesis directly controls the vectors separated by the
detection loss. Generating nodes or graphs also specifies the
attributes and connections presented to the encoder. The two
routes offer different control over a negative example: its
position in representation space or its observable graph content.

Generated graphs can train a detector or shape its reference prototypes. \textbf{PGR-MOOD}~\cite{PGR-MOOD} perturbs a pretrained diffusion model's score-function parameters to generate OOD proxies. Attraction toward IND graphs and repulsion from these proxies guide prototype generation. Test graphs are matched to these prototypes using Fused Gromov--Wasserstein similarity, which compares node attributes and structural relationships. The prototypes replace the per-input reconstruction used by the paper's GR-MOOD baseline. \textbf{PGOS}~\cite{sun2026pgos} learns a reinforcement-learning policy to explore informative low-density latent regions, then decodes selected representations into pseudo-OOD graphs for detector training.

Semantic control specifies which changes should trigger rejection. \textbf{GDDA}~\cite{he2025gdda} addresses semantic OOD detection under covariate shift. It separates semantic factors from domain-specific styles and uses controlled diffusion to generate pseudo-IND and pseudo-OOD representations. These train an energy-based detector to reject unknown classes while accepting shifted known-class graphs. \textbf{SA-Diff}~\cite{dong2026sadiff} perturbs semantic conditioning in a class-conditional graph diffusion model to generate structurally valid graphs with shifted semantics. Its outliers support boundary-aware fine-tuning and a test-time score comparing $k$-nearest-neighbor distances to IND and generated OOD references.

Language models provide semantic guidance for node generation. \textbf{LLMGuard}~\cite{lv2025llmguard} generates challenging text-attributed nodes with an LLM and attaches them to the graph using lightweight edge predictors. \textbf{LECT}~\cite{ma2026lect} generates dependency-aware pseudo-OOD nodes and separates them from IND nodes through energy contrastive learning. Both construct unfamiliar content together with the relational context needed to use it during detection training.

\subsection{Meta-learning}
\label{sec:outlier_episodic}

\textbf{SMUG}~\cite{SMUG} trains unknown-class rejection through meta-learning for few-shot node classification. Selected observed classes act as artificial OOD classes in training episodes, providing labels for OOD discriminators. The encoder is optimized jointly for known-class prediction and unobserved-class detection. Varying the known classes across tasks exposes the model to different rejection boundaries.

\noindent\textbf{Discussion.}
The value of outlier supervision depends on its correspondence to the rejection target. Semantic novelty, domain changes, and scene-background anomalies require different training contrasts. Comparisons should identify how auxiliary examples or labels are obtained and how their semantics and construction relate to the test OOD data.

\section{Detection Calibration}
\label{sec:C}
\label{sec:predictive_calibration}

\begin{figure}[t]
\centering
\includegraphics[width=\columnwidth]{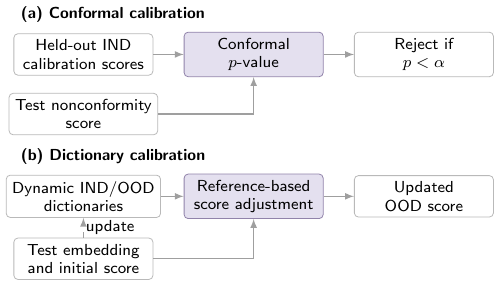}
\caption{Detection calibration using conformal comparison or reference dictionaries.}
\label{fig:calibration_framework}
\end{figure}

Calibration uses reference data to adjust OOD scores or select a rejection threshold (Fig.~\ref{fig:calibration_framework}). Graph-level methods use held-out IND graphs for conformal inference or construct reference dictionaries from test graphs.
Graph conformal prediction establishes coverage under random
transductive splits and suitable permutation-invariance conditions,
providing a basis for studying calibration with dependent node
observations~\cite{huang2023cfgnn}.

\subsection{Conformal Calibration}
\label{sec:calibration_conformal}

\textbf{CGOD}~\cite{lin2025cgod} combines aggregated nonconformity scores with split conformal inference. Score consistency and representation diversity guide learned augmentations in graph-representation space. Scores from the augmented views are aggregated for detection. One IND subset trains the augmentations; a separate, held-out subset supplies the calibration scores used to calculate a test graph's conformal $p$-value.
With exchangeable IND scores and a scoring rule learned independently of the held-out calibration and test graphs, rejecting graphs whose $p$-values fall below $\alpha$ bounds the marginal IND false-rejection probability by $\alpha$. The data split connects the rejection threshold to a specified error rate while allowing the nonconformity score to be learned.

\subsection{Dictionary-based Calibration}
\label{sec:calibration_dictionary}

\textbf{D2GO}~\cite{hou2025d2go} estimates graphons from test graphs and uses graphon mix-up to generate boundary references. Pseudo-labels and priority queues maintain dynamic IND and OOD dictionaries. Similarities to these dictionaries recalibrate the original OOD scores without fine-tuning the backbone.
\textbf{BaCa}~\cite{hou2026baca} uses boundary-aware topology generation and attention-guided dual dynamic dictionaries. The pretrained model remains fixed while test graphs provide evolving references. Generated and pseudo-labeled samples serve as calibration references, with their assigned IND/OOD identities and dictionary coverage affecting the resulting scores.

\noindent\textbf{Discussion.}
Conformal inference and dictionary-based recalibration modify
different parts of the detection decision. CGOD ties rejection
to a marginal IND false-rejection budget using a score learned
independently of the held-out reference set. Dictionary methods
alter sample comparisons by updating their references with
pseudo-labeled test graphs, making test composition and update
rules part of the scoring mechanism. Updating the dictionaries
changes the score function even with a frozen encoder. A
false-rejection guarantee for an adaptive detector must account
for these updates and the construction of calibration scores.
OOD rejection at the selected threshold measures the separation
achieved by either approach.

\section{Datasets, Benchmarks, and Evaluation}
\label{sec:benchmarks}

\begin{table*}[!t]
\centering
\caption{Representative graph datasets and OOD detection settings.}
\label{tab:good_dataset_families}
\begingroup
\footnotesize
\setlength{\tabcolsep}{4pt}
\renewcommand{\arraystretch}{1.14}
\begin{tabularx}{\textwidth}{@{}p{0.75cm}p{2.35cm}>{\raggedright\arraybackslash}X>{\raggedright\arraybackslash}X@{}}
\toprule
Unit & Domain & Representative sources & OOD Settings in Representative Studies\\
\midrule
Node & Academic / collaboration & Cora, CoraML, Citeseer, PubMed, DBLP, Coauthor-CS/Physics, ogbn-arxiv & Class holdout, feature/structure interventions, temporal partition~\cite{wu2023gnnsafe,GRASP}; multi-label settings~\cite{cai2025mlgood}.\\
Node & Co-purchase & Amazon-Computers, Amazon-Photo & Unknown classes and feature/structure changes~\cite{wu2023gnnsafe}.\\
Node & Social networks & Twitch, Reddit & Regional partition or unknown-class construction~\cite{wu2023gnnsafe,GRASP}.\\
Node & Biological / multi-label & ogbn-proteins, PPI, PCG, HumLoc, EukLoc & Feature-based and cross-dataset detection with multi-label targets~\cite{cai2025mlgood}.\\
\midrule
Graph & Molecular & BZR, COX2, AIDS, DHFR; Tox21, SIDER, BBBP, BACE, ESOL, MUV & Cross-dataset detection~\cite{GOOD-D}; molecular domain settings~\cite{PGR-MOOD}.\\
Graph & Protein / enzyme & ENZYMES, PROTEINS & Cross-dataset graph detection~\cite{GOOD-D}.\\
Graph & Social interaction & IMDB-BINARY, IMDB-MULTI & Cross-dataset graph detection~\cite{GOOD-D}.\\
\bottomrule
\end{tabularx}
\endgroup
\end{table*}


In a class-holdout benchmark, an observation is OOD if its class was withheld from training. When the OOD source is a different graph collection, attributes, connectivity, and class composition can change together. Our synthetic node benchmark varies feature covariance, edge structure, and class prototypes separately to identify which changes each detector detects.

\subsection{Datasets for GOOD Detection}
\label{sec:benchmark_datasets}

Table~\ref{tab:good_dataset_families} groups datasets by detection target. Node-level studies score individual nodes in citation, co-authorship, social, and biological networks.
In ML-GOOD, a node can carry multiple class labels~\cite{cai2025mlgood}.

Cross-dataset benchmarks draw IND graphs from one collection and OOD graphs from another.
These evaluations use molecular, biological, and social graph collections from resources such as TUDataset and MoleculeNet~\cite{morris2020tudataset,wu2018moleculenet,GOOD-D}.

\subsection{OOD Data Construction}
\label{sec:benchmark_shifts}

Class holdout treats observations from unseen classes as OOD. Feature perturbations change node attributes, and structural perturbations change graph connectivity.
Domain and temporal splits assign OOD status by source or collection period, so attributes, topology, and class composition can change together.
On image benchmarks, Wang et al.~\cite{wang2025dissecting} evaluate semantic and covariate shifts separately and show that outlier-exposure performance depends on the similarity between auxiliary outliers and test OOD data.

\subsection{GOOD Detection Benchmarks}
\label{sec:benchmark_resources}

\begin{table*}[!t]
\centering
\caption{Comparison of representative graph OOD detection benchmarks.}
\label{tab:benchmark_scope}
\begingroup
\footnotesize
\setlength{\tabcolsep}{4pt}
\renewcommand{\arraystretch}{1.14}
\begin{tabularx}{\textwidth}{@{}p{2.0cm}p{0.8cm}>{\raggedright\arraybackslash}X>{\raggedright\arraybackslash}X>{\raggedright\arraybackslash}X@{}}
\toprule
Benchmark & Level & Training Setting & OOD Settings & Evaluation Focus\\
\midrule

GNNSafe~\cite{wu2023gnnsafe}
& Node
& Labeled IND data; optional outlier exposure
& Feature and structural shifts, class holdout, domain and temporal shifts
& Detection across different distribution shifts and training settings\\

GOOD-D~\cite{GOOD-D}
& Graph
& Unlabeled IND graphs
& Cross-dataset IND/OOD graph pairs
& Unsupervised graph-level OOD detection\\

UB-GOLD~\cite{wang2025ubgold}
& Graph
& Unsupervised graph-level detection
& OOD detection and graph anomaly detection across 35 datasets
& Detection performance, robustness, and computational efficiency\\

\midrule

Synthetic Benchmark (Ours)
& Node
& IND training, validation, and calibration; method-specific training resources
& Controlled feature, structural, semantic, and compound shifts
& Detection sensitivity to distribution shifts and graph structure\\

\bottomrule
\end{tabularx}
\endgroup
\end{table*}
GNNSafe~\cite{wu2023gnnsafe} evaluates node detection under class holdout, feature and structural perturbations, and temporal or regional shifts.
GOOD-D~\cite{GOOD-D} evaluates unsupervised graph detection on ten IND/OOD dataset pairs.
UB-GOLD~\cite{wang2025ubgold} compares 18 methods for graph-level OOD and anomaly detection across 35 datasets and four scenarios.
Table~\ref{tab:benchmark_scope} compares their evaluation settings and coverage.

GOOD~\cite{gui2022goodbenchmark} and DrugOOD~\cite{ji2023drugood} define distribution shifts for evaluating predictive generalization on graphs.
Detection studies use such splits to form IND and OOD evaluation sets~\cite{PGR-MOOD,wang2025ubgold}.
GOOD-D and UB-GOLD include graph-kernel baselines that score outliers using structural similarity~\cite{GOOD-D,wang2025ubgold}.
These baselines pair Weisfeiler--Lehman or propagation kernels with conventional detectors such as one-class SVM, isolation forest, or local outlier factor~\cite{neumann2016propagation,scholkopf2001ocsvm,liu2008isolation,breunig2000lof}.

\subsection{Evaluation Metrics}
\label{sec:benchmark_metrics}

GOOD detection performance is commonly evaluated using AUROC, AUPR, and FPR95. Let $S$ denote an OOD detection score, with larger values indicating stronger OOD evidence. 
Let $S$ be the detection score. A node or graph is when $S>\tau$.
The OOD recall and the fraction of IND observations flagged as OOD are given by
\begin{equation}
\begin{aligned}
\mathrm{TPR}_{\mathrm{OOD}}(\tau)&=\Pr(S>\tau\mid\mathrm{OOD}),\\
\mathrm{FPR}_{\mathrm{IND}}(\tau)&=\Pr(S>\tau\mid\mathrm{IND}).
\end{aligned}
\label{eq:good_eval_rates}
\end{equation}
For independently drawn OOD and IND observations with scores $S^+$ and $S^-$,
AUROC is the probability that the OOD observation scores higher, with half credit for ties.
\begin{equation}
\mathrm{AUROC}
=\Pr(S^+>S^-)+\tfrac12\Pr(S^+=S^-).
\label{eq:good_eval_auc}
\end{equation}
We report AUPR as noninterpolated average precision (AP), with OOD as the positive class:
\begin{equation}
\mathrm{AUPR}=\mathrm{AP}
=\sum_{j=1}^{J}(R_j-R_{j-1})P_j.
\label{eq:good_eval_ap}
\end{equation}
Here $P_j$ and $R_j$ are precision and recall after including all observations at the $j$th distinct score in descending order, with $R_0=0$.
We report AP separately at each OOD prevalence because precision depends on the proportion of OOD observations.

Our FPR95 is the fraction of OOD observations accepted as IND when the threshold is set to the 95th percentile of IND scores.
Let $F_{\mathrm{IND}}(t)=\Pr(S\leq t\mid\mathrm{IND})$, then:
\begin{equation}
\begin{aligned}
\tau_{\mathrm{IND},95}
 &=\inf\{t:F_{\mathrm{IND}}(t)\geq0.95\},\\
\mathrm{FPR95}
 &=\Pr(S\leq\tau_{\mathrm{IND},95}\mid\mathrm{OOD}).
\end{aligned}
\label{eq:good_eval_fpr95}
\end{equation}
Studies that fix OOD recall at 95\% report the fraction of IND observations rejected as OOD at that operating point.

\subsection{Synthetic GOOD Detection Benchmark}
\label{sec:controlled_good}

The benchmark uses a contextual stochastic block model (CSBM)~\cite{deshpande2018contextual}.
Within each connectivity regime, we vary neighborhood relations while holding node degrees and class-pair edge counts fixed.
The same model lets us vary feature covariance or class prototypes on a fixed graph.

\subsubsection{Graph Model and Detection Target}

We generate graphs with $n=1200$ nodes, three balanced classes, 16-dimensional node features, and an expected degree of 12.
Each independently generated dataset has orthonormal class directions $\{e_k\}_{k=1}^{3}$, novel-class directions $\{f_k\}_{k=1}^{3}$,
a latent role direction $v$, and nuisance directions $u_1,u_2$. Node features are generated as:
\begin{equation}
\begin{aligned}
x_i &= 3e_{c_i}+r_i v+\epsilon_i,
\quad r_i\sim\operatorname{Unif}\{-1,1\},\\
\epsilon_i &\sim\mathcal N(0,I_{16}+3u_1u_1^\top),
\end{aligned}
\label{eq:csgood_features}
\end{equation}
where \(c_i\) is node \(i\)'s original class, and \(r_i\) is the latent role used in the structural intervention.

We sample edges independently for unordered pairs of distinct nodes using the following probabilities~\cite{zheng2026graph}:
\begin{equation}
\Pr(A_{ij}=1\mid c_i,c_j)=
\begin{cases}
\dfrac{12h}{n/3-1}, & c_i=c_j,\\[4pt]
\dfrac{12(1-h)}{n-n/3}, & c_i\ne c_j.
\end{cases}
\label{eq:csgood_edges}
\end{equation}
The parameter \(h\) sets the fraction of the expected degree contributed by same-class neighbors.
We use \(h=1/6\), \(399/1199\), and \(2/3\) for heterophilic, neutral, and homophilic connectivity, respectively.
At the neutral value, within-class and between-class node pairs have the same edge probability.

Independently of the original graph, we select OOD-source nodes uniformly at prevalence $\rho\in\{0.1,0.5\}$.
The detection target is source membership \(m_i\), with $m_i=1$ or a selected node and \(m_i=0\) otherwise.

\subsubsection{Controlled OOD Shifts}

We construct four OOD settings by modifying different aspects of the generating distribution.
The main comparisons use the strongest predefined intervention level for each shift type, with 50\% OOD nodes in the test graph.

\noindent\textbf{Feature Shift (F).}
We rotate the nuisance feature directions $(u_1,u_2)$ of OOD-source nodes by $\theta\in\{\pi/12,\pi/6,\pi/4\}$. This changes feature covariance while preserving graph connectivity, individual feature norms, and the population class posterior given raw node features. The resulting shift evaluates detection sensitivity to class-preserving changes in the feature distribution.

\noindent\textbf{Structural Shift (S).}
We introduce relational changes through constrained edge rewiring. The structural intervention is governed by
\begin{equation}
T(A)=\sum_{(i,j)\in E}
\max(m_i,m_j)r_i r_j,
\label{eq:csgood_structure}
\end{equation}
where an eligible edge swap $A\rightarrow A'$ is accepted with probability
$\min\{1,\exp[\lambda(T(A')-T(A))]\}$,
for $\lambda\in\{0.25,0.5,1\}$.
Across \(5\lvert E\rvert\) swap attempts, we hold node features, degrees, class-pair edge counts,
cross-source edge counts, and connected components fixed.
Rewiring can change the neighborhoods of both OOD-source nodes and their IND neighbors.

\noindent\textbf{Open-set Semantic Shift (O).}
The detection target is a node from an unseen class~\cite{OpenWGL,song2022OODGAT,LMN}.
We generate novel class prototypes by rotating each original direction \(e_k\) toward an orthogonal direction \(f_k\):
$3(\cos\phi\,e_k+\sin\phi\,f_k)$,
where $\phi\in\{\pi/12,\pi/6,\pi/3\}$.
Each novel prototype has norm 3. Graph connectivity and the class-conditional noise distribution stay fixed.
At the task level, this corresponds to GNNSafe's label leave-out setting, which treats held-out classes as OOD~\cite{wu2023gnnsafe},
but the controlled prototype construction and independent-graph access protocol are different.

\noindent\textbf{Compound Shift (F+S).}
F+S uses the same OOD-source assignment for feature rotation and constrained edge rewiring.
The full-null control tests whether detector scores separate the two source groups in the absence of an intervention.
The resulting benchmark supports controlled comparisons of individual and combined distribution shifts.

\subsubsection{Detection Evaluation}

We evaluate six representative GOOD detectors covering five of the six
categories in our taxonomy: TIDE~\cite{wang2026tide} for representation
shaping; DeGEM~\cite{anonymous2025DeGEM} and GPN~\cite{GPN} for IND
distribution modeling; NODESAFE~\cite{NODESAFE} for predictive evidence
modeling; GNNSafe~\cite{wu2023gnnsafe} for relational evidence modeling;
and GOLD~\cite{wang2025GOLD} for IND--OOD boundary learning. Maximum
Softmax Probability (MSP)~\cite{hendrycks2017baseline} and
Energy~\cite{liu2020energy} serve as conventional OOD scoring baselines.

The evaluation uses ten independently generated datasets for each
connectivity setting: heterophilic, neutral, and homophilic. Each
dataset provides separate IND graphs for training, validation, and
calibration. Detectors assign OOD scores to nodes in an unlabeled test
graph. MSP, Energy, and GNNSafe share the same trained backbone, while
the remaining detectors are trained separately. GOLD uses internally
generated outlier embeddings during training. Detection performance
is measured by AUROC, AUPR, and FPR95. IND classification accuracy is
reported where applicable.

D1 measures how typical a node's features are under the IND training distribution.
We fit a Gaussian mixture with one mean per class and a shared covariance matrix,
and score each node by the negative log density of its features under this mixture:
\begin{equation}
    S_{\mathrm{D1}}(x)
    =-\log\sum_{k=1}^{3}
    \widehat\pi_k
    \mathcal N(x;\widehat\mu_k,\widehat\Sigma).
    \label{eq:csgood_d1}
\end{equation}
D2 assigns high scores to nodes whose agreement with their neighbors is unusually high or low relative to IND nodes.
For a node with neighbors, \(q_i\) is the mean inner product between its features and theirs.
We set \(q_i=0\) for an isolated node. 
 \begin{equation}
    \begin{aligned}
    q_i&=\frac{x_i^\top\sum_{j\in\mathcal N(i)}x_j}
    {\max(d_i,1)},\\
    S_{\mathrm{D2}}(i)
    &=-\log\!\left[
    2\min\{\widehat F_q(q_i),1-\widehat F_q(q_i)\}
    \right],
    \end{aligned}
    \label{eq:csgood_d2}
\end{equation}
where $\widehat F_q$ is the empirical cumulative distribution function of $q$ for IND nodes.

Three reference scores use information from the generator.
A likelihood ratio targets the change in feature covariance, a projection onto the novel class directions targets semantic shift,
and a role-based neighborhood statistic targets structural shift.
Comparing these scores with D1 and D2 shows which source differences are visible in node features or neighborhood agreement estimated from IND data.



\section{Method Capabilities and Comparative Analysis}
\label{sec:good_comparison}

GOOD detectors differ in the information they require and the
distributional changes they can identify. We compare representative
methods by their detection mechanisms and data requirements, and
use controlled experiments to examine detector sensitivity and
rejection performance under different shifts.

\subsection{Detection Mechanisms and Information Requirements}
\label{sec:comparison_capabilities}

Table~\ref{tab:cross_method_capability} compares 18 representative
methods across the six categories, listing their detection mechanisms
and information requirements.

\begin{table*}[!t]
    \centering
    \caption{Detection mechanisms and information requirements of representative GOOD methods.}
    \label{tab:cross_method_capability}
    \begingroup
    \footnotesize
    \setlength{\tabcolsep}{4pt}
    \renewcommand{\arraystretch}{1.15}
    
    \begin{tabularx}{\textwidth}{@{}
        >{\raggedright\arraybackslash}p{2.25cm}
        c
        >{\raggedright\arraybackslash}p{5.6cm}
        >{\raggedright\arraybackslash}X@{}}
    \toprule
    Method & Unit & Detection mechanism & Information used \\
    \midrule
    
    \multicolumn{4}{@{}l}{\textbf{Representation Shaping}}\\[1pt]
    
    GOOD-D~\cite{GOOD-D} & G
    & Contrastive consistency across node, graph, and group representations
    & Unlabeled IND training graphs \\
    
    TNT-OOD~\cite{wang2025tntood} & N
    & Alignment of textual and topological representations
    & IND node labels; node text and pretrained text representations \\
    
    UniGOOD~\cite{liao2026unigood} & G
    & Spectral contrastive learning of invariant subgraphs
    & Labeled IND graphs; an unlabeled graph pool \\
    
    \midrule
    \multicolumn{4}{@{}l}{\textbf{IND Distribution Modeling}}\\[1pt]
    
    GraphDE~\cite{li2022GraphDE} & G
    & Generative modeling with latent IND/OOD membership
    & Labeled training graphs, potentially containing outliers \\
    
    GPN~\cite{GPN} & N
    & Class-conditional latent densities and posterior propagation
    & IND node labels; graph context for posterior propagation \\
    
    EnergyDef~\cite{EnergyDef2024kdd} & N
    & Energy-based density estimation
    & Known-IND training nodes; optional OOD supervision;
      OOD validation for model selection in reported settings \\
    
    DeGEM~\cite{anonymous2025DeGEM} & N
    & Latent-space energy modeling
    & IND node labels; internally sampled latent negatives \\
    
    \midrule
    \multicolumn{4}{@{}l}{\textbf{Predictive Evidence Modeling}}\\[1pt]
    
    GKDE~\cite{GKDE} & N
    & Dirichlet evidence and vacuity-based detection
    & IND node labels and graph distances \\
    
    ML-GOOD~\cite{cai2025mlgood} & N
    & Aggregation of label-wise energy scores
    & Multi-label IND annotations; auxiliary OOD training nodes
      in the evaluated setting \\
    
    GLIP-OOD~\cite{xu2025glipood} & N
    & Graph--text semantic matching
    & Pretrained GraphCLIP; IND class names;
      OOD class names or LLM-generated alternatives \\
    
    \midrule
    \multicolumn{4}{@{}l}{\textbf{Relational Evidence Modeling}}\\[1pt]
    
    GNNSafe~\cite{wu2023gnnsafe} & N
    & Propagation of energy scores
    & IND node labels; auxiliary outliers in exposure variants \\
    
    GRASP~\cite{GRASP} & N
    & Score propagation with IND reference nodes
    & Pretrained IND classifier; known-IND nodes available at inference \\
    
    RAGNOR~\cite{gu2025ragnor} & N/G
    & Robust neighborhood comparison of embedding norms
    & Pretrained backbone; IND norm statistics; test-graph neighborhoods \\
    
    \midrule
    \multicolumn{4}{@{}l}{\textbf{IND--OOD Boundary Learning}}\\[1pt]
    
    HGOE~\cite{HGOE2024mm} & G
    & Discriminative learning with external and synthetic outliers
    & IND training graphs; external outliers; generated graphs \\
    
    SMUG~\cite{SMUG} & N
    & Meta-learning with artificial unknown classes
    & Meta-training class labels; few-shot support labels \\
    
    NOAOM~\cite{zhou2026noaom} & G
    & Prototype-guided synthesis of near-OOD embeddings
    & IND training graphs; generated near-OOD embeddings \\
    
    \midrule
    \multicolumn{4}{@{}l}{\textbf{Detection Calibration}}\\[1pt]
    
    CGOD~\cite{lin2025cgod} & G
    & Aggregated nonconformity scores with conformal calibration
    & Pretrained detector; separate IND sets for augmentation learning
      and calibration \\
    
    D2GO~\cite{hou2025d2go} & G
    & Score calibration with dynamic reference dictionaries
    & Pretrained detector; unlabeled test graphs for dictionary updates \\
    
    \bottomrule
    \end{tabularx}
    
    \par\smallskip
    {\footnotesize
    N: node-level detection; G: graph-level detection; N/G: both.}
    \endgroup
    \end{table*}

DeGEM fits a latent energy model with internally sampled negatives;
HGOE uses external and synthetic outliers to learn an IND--OOD
boundary~\cite{anonymous2025DeGEM,HGOE2024mm}. EnergyDef fits its
energy model without IND class labels and uses OOD validation data
for model selection in the reported
settings~\cite{EnergyDef2024kdd}.

GPN propagates evidence from class-conditional latent densities to
form node posteriors, whereas GNNSafe propagates energy scores
computed from classification logits~\cite{GPN,wu2023gnnsafe}.
GRASP adds known-IND reference nodes to the propagation
graph~\cite{GRASP}.

TNT-OOD learns text--topology alignment from labeled IND
nodes~\cite{wang2025tntood}. GLIP-OOD scores graph--text compatibility
with a pretrained graph foundation model and class names, without
task-specific node annotations. Its main variant generates pseudo-OOD
class names from unlabeled node text using an
LLM~\cite{xu2025glipood}.

CGOD learns augmentations on one IND subset and uses a separate
subset for conformal calibration. Its marginal IND false-rejection
guarantee requires a scoring rule learned independently of the
calibration and test graphs, with exchangeable IND calibration and
test samples~\cite{lin2025cgod}. D2GO updates graphon dictionaries
from unlabeled test graphs and uses them to recalibrate existing
scores~\cite{hou2025d2go}.

\subsection{GOOD Detection under Controlled Distribution Shifts}
\label{sec:controlled_results}
\label{sec:comparison_evidence}

\begin{figure*}[t]
    \centering
    \includegraphics[width=.98\textwidth]{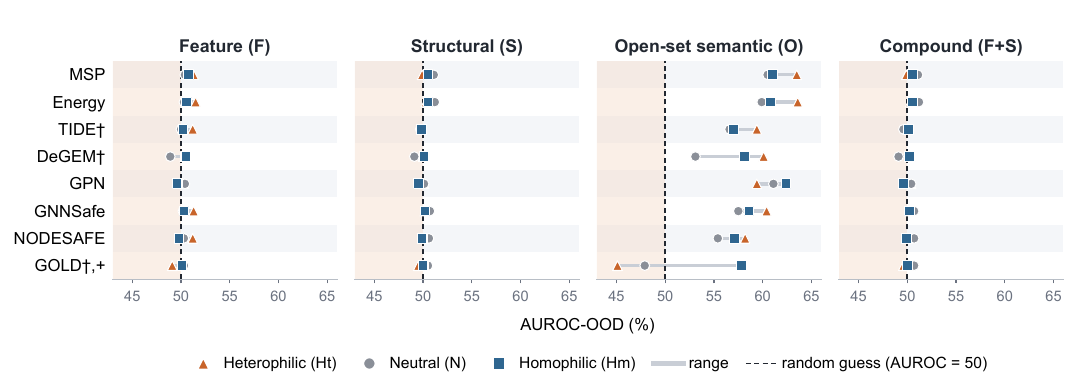}
    \caption{OOD detection AUROC (\%) under feature (F), structural (S), open-set semantic (O), and compound (F+S) shifts across heterophilic (Ht), neutral (N), and homophilic (Hm) graphs.}
    \label{fig:controlled_detector_matrix}
\end{figure*}

The mean AUROCs in Fig.~\ref{fig:controlled_detector_matrix} range
from 48.9\% to 51.5\% for F, S, and their combination F+S across
the eight evaluated methods and three graph environments. O yields
a wider range, from 45.1\% to 63.6\%.

Under O, GPN obtains 59.4\% AUROC on heterophilic graphs and 62.4\%
on homophilic graphs; Energy obtains 63.6\% and 60.8\% in the same
two environments. These values come from models trained separately
in each graph environment. With the same trained backbone on neutral
graphs under O, GNNSafe's score propagation changes AUROC from
Energy's 59.9\% to 57.5\%.

\begin{figure*}[t]
    \centering
    \includegraphics[width=.95\textwidth]{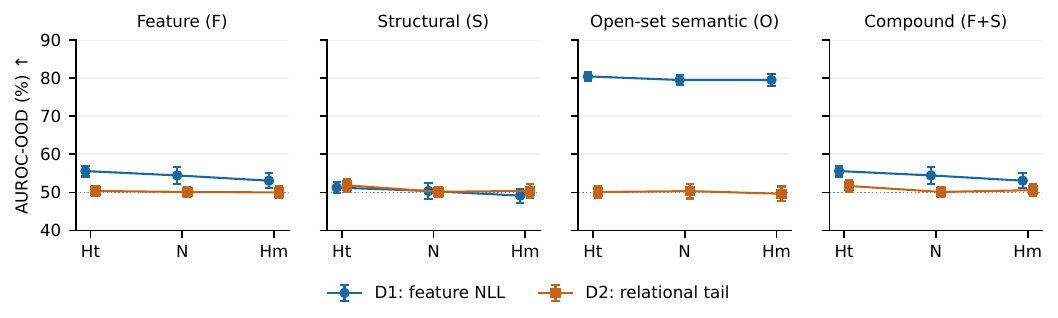}
    \caption{AUROC (\%) of the IND-based feature-density (D1) and neighborhood-agreement (D2) scores across the four shifts and three graph environments. Points show means and error bars show standard deviations across ten independently generated datasets. The dotted line marks 50\% AUROC.}
    \label{fig:controlled_access}
\end{figure*}

D1 fits a class-conditional feature-density model to labeled IND
training data. D2 scores unusually low or high feature--neighborhood
agreement relative to IND training graphs
(Fig.~\ref{fig:controlled_access}). Under O, D1 reaches approximately
80\% AUROC across the three graph environments. Its AUROC on neutral
graphs is 79.5\%, compared with 61.1\% for GPN. D1 has lower AUROCs
under F and F+S and remains near chance under S, which preserves node
features. D2 remains near chance under S, giving little IND--OOD
separation with the chosen neighborhood-agreement score.

\begin{table*}[!t]
\centering\footnotesize
\setlength{\tabcolsep}{3pt}
\renewcommand{\arraystretch}{1.12}
\caption{OOD detection performance in terms of FPR95 ($\downarrow$) across distribution shifts and graph environments. Ht/N/Hm denote heterophilic, neutral, and homophilic graphs.}
\label{tab:controlled_fpr95}
\label{tab:controlled_semantic}
\begin{tabular}{ll*{8}{c}}
\toprule
Shift & Env. & MSP & Energy & TIDE$^{\dagger}$ & DeGEM$^{\dagger}$ & GPN & GNNSafe & NODESAFE & GOLD$^{\dagger,+}$ \\
\midrule
F & Ht & $94.5\pm1.4$ & $94.2\pm1.1$ & $94.1\pm1.4$ & $95.8\pm1.1$ & $94.7\pm1.8$ & $94.6\pm1.1$ & $94.2\pm1.0$ & $95.7\pm0.8$ \\
F & N & $94.1\pm1.5$ & $94.9\pm1.3$ & $94.9\pm1.3$ & $95.1\pm0.9$ & $94.5\pm1.6$ & $95.0\pm1.3$ & $94.9\pm1.8$ & $94.7\pm1.1$ \\
F & Hm & $94.2\pm1.3$ & $94.5\pm1.3$ & $94.5\pm1.2$ & $95.1\pm1.0$ & $94.5\pm1.4$ & $94.3\pm1.4$ & $94.5\pm0.8$ & $94.4\pm1.4$ \\
\midrule
S & Ht & $94.9\pm0.9$ & $94.3\pm1.3$ & $94.3\pm1.1$ & $94.2\pm1.6$ & $94.7\pm1.9$ & $94.2\pm1.2$ & $94.4\pm1.5$ & $95.2\pm1.0$ \\
S & N & $94.2\pm1.4$ & $94.0\pm1.7$ & $94.9\pm1.7$ & $94.9\pm1.2$ & $94.7\pm1.4$ & $93.9\pm2.3$ & $94.7\pm1.3$ & $94.2\pm1.9$ \\
S & Hm & $94.7\pm1.1$ & $94.5\pm1.3$ & $94.8\pm1.2$ & $94.5\pm1.5$ & $94.7\pm1.6$ & $95.2\pm1.4$ & $94.8\pm0.9$ & $94.7\pm1.1$ \\
\midrule
O & Ht & $89.7\pm1.4$ & $91.5\pm2.4$ & $91.6\pm2.2$ & $87.6\pm7.7$ & $92.2\pm2.3$ & $91.6\pm1.5$ & $92.5\pm2.1$ & $96.9\pm0.9$ \\
O & N & $90.4\pm1.8$ & $92.5\pm1.3$ & $92.4\pm1.6$ & $95.2\pm5.8$ & $91.6\pm2.1$ & $92.7\pm1.3$ & $93.8\pm1.4$ & $96.6\pm1.5$ \\
O & Hm & $89.0\pm1.7$ & $90.3\pm1.5$ & $91.0\pm1.5$ & $89.3\pm9.5$ & $90.8\pm1.7$ & $90.4\pm2.7$ & $91.6\pm1.7$ & $91.6\pm1.9$ \\
\midrule
F+S & Ht & $95.2\pm0.9$ & $94.3\pm1.4$ & $94.5\pm0.9$ & $94.3\pm1.5$ & $94.7\pm1.9$ & $94.3\pm1.1$ & $94.4\pm1.6$ & $95.0\pm1.0$ \\
F+S & N & $94.0\pm1.3$ & $94.2\pm1.9$ & $94.9\pm1.8$ & $94.7\pm1.3$ & $94.5\pm1.2$ & $94.0\pm2.4$ & $94.7\pm1.3$ & $94.2\pm1.9$ \\
F+S & Hm & $94.9\pm1.1$ & $94.6\pm1.2$ & $94.9\pm1.2$ & $94.5\pm1.3$ & $94.7\pm1.6$ & $95.2\pm1.3$ & $94.7\pm1.3$ & $94.8\pm1.1$ \\
\bottomrule
\end{tabular}
\par\smallskip
\begin{minipage}{\textwidth}\footnotesize
$\dagger$ indicates released-code variants; $+$ indicates synthetic OOD exposure in GOLD.
\end{minipage}
\end{table*}

Table~\ref{tab:controlled_fpr95} reports OOD acceptance when the
threshold is chosen from IND test scores to retain at least 95\%
of IND test nodes. Under F, S, and F+S, approximately 94\%--96\% of
OOD nodes are accepted. Under O on neutral graphs, GPN obtains
61.1\% AUROC and accepts 91.6\% of OOD nodes at this threshold.

On homophilic graphs under O, the classifier shared by MSP, Energy,
and GNNSafe reaches 98.4\% accuracy on IND-source nodes. The
corresponding OOD AUROCs are 61.0\%, 60.8\%, and 58.6\%, respectively.





\subsection{Comparative Insights}
\label{sec:comparison_critical}
%
A detector's sensitivity depends on the information retained in its
representation and the way that information is scored. An invariance
that supports known-class prediction can suppress a class-preserving
change that the detection task treats as OOD. Substructure encoding
and information decomposition make different choices about which graph
information to retain~\cite{SGOOD,wang2026tide}. Latent densities,
predictive uncertainty, and semantic matching turn the retained
information into detection scores. In the controlled semantic
comparison, the IND-fitted D1 score separates the sources more strongly
than GPN, showing that raw IND feature statistics provide an effective
reference for this shift (Fig.~\ref{fig:controlled_access}).

The value of graph context depends on the evidence being combined and
the IND/OOD memberships of connected nodes. Propagation can reinforce
consistent evidence or mix scores across source boundaries. Neighborhood
comparisons depend on whether local differences reflect the intended
OOD change~\cite{wu2023gnnsafe,GRASP,TopoOOD}. Class homophily describes
a different relation: nodes from different known classes can both be
IND. Initial score quality and local IND/OOD mixing help determine
which neighbors provide useful detection information. The controlled
structural interventions leave several graph-based scores near
chance, illustrating how a detector can use connectivity while
responding weakly to the induced change in feature--edge dependence
(Fig.~\ref{fig:controlled_detector_matrix}).

Auxiliary samples contribute through model fitting, rejection
supervision, or reference construction. Negatives sampled from an
energy model approximate the model expectations used for
fitting~\cite{anonymous2025DeGEM}. Outliers used for boundary learning
specify which departures receive rejection supervision; their validity,
diversity, and relation to the test OOD sources shape the learned
distinction~\cite{HGOE2024mm}. Calibration reference sets determine
the comparisons made at inference. Their selection and update rules
affect the resulting scores even when the backbone remains
fixed~\cite{hou2025d2go,hou2026baca}.

Available labels and reference data guide the choice of detector.
Unlabeled IND graph collections support contrastive training, while
semantic matching uses class
descriptions and pretrained graph--text
representations~\cite{GOOD-D,xu2025glipood}. Post-hoc scoring and
relational comparison are options for frozen models when the required
statistics and graph context are available~\cite{gu2025ragnor}.
Method selection needs to account for the information used in model
fitting, model selection, calibration, and inference, including access
to unlabeled test graphs.

Score ranking and threshold validity describe different properties of
a detector. The controlled results show that higher AUROC can
coexist with substantial OOD acceptance at high IND retention.
Conformal calibration controls marginal IND false rejection under
its sampling and scoring assumptions~\cite{lin2025cgod}; detection power
depends on score separation at the chosen threshold. When adaptation or
reference updates change the scoring rule, thresholds need to be
assessed under the updated rule. Comparisons should report OOD
rejection at the required IND retention rate and identify the data
used to choose the threshold.

\section{Future Directions and Conclusion}
\label{sec:future_directions}

Progress in GOOD detection depends on preserving the changes that define OOD inputs and turning semantic, attribute, and relational information into reliable rejection decisions.

\subsection{Future Directions}

\noindent\textbf{Detection-Oriented Representations.}
Detection objectives should specify which semantic, attribute, and relational changes must remain visible in a representation. Semantic OOD detection under covariate shift illustrates this requirement: known-class graphs remain acceptable as their domain characteristics change~\cite{he2025gdda}. Substructure encoding and information decomposition provide tools for preserving complementary graph information~\cite{SGOOD,wang2026tide}. Future work could learn separate representations of class semantics, feature statistics, and attribute--edge dependence, with augmentation and compression guided by the rejection target. Extending information-theoretic feature construction~\cite{mondal2025variational} and OOD learnability analysis~\cite{fang2024learnability} to graph data could clarify which shifts remain detectable after node encoding and graph pooling.

\noindent\textbf{Reliable Use of Graph Context.}
Graph context should be selected according to its contribution to a node's IND/OOD decision. Uncertainty modeling, score propagation, and reference augmentation provide different ways to incorporate neighboring information~\cite{GPN,wu2023gnnsafe,GRASP}. One route is to estimate IND compatibility separately from a node's attributes and its neighborhood. Agreement and uncertainty in these estimates could guide neighbor selection and propagation depth, with known-IND reference nodes anchoring the comparison. Analysis should connect these choices to initial score quality and local IND/OOD mixing, with class homophily treated as a separate graph property.

\noindent\textbf{Outlier Supervision with Broader Coverage.}
Outlier supervision should provide broad coverage of plausible OOD changes and informative examples near the rejection boundary. External exposure and synthetic outliers already support boundary learning~\cite{HGOE2024mm,wang2025GOLD,zhou2026noaom}. Future generators could vary semantic content, node attributes, and connectivity while enforcing graph validity and an explicit departure from the task's IND scope. For latent outliers, linking perturbations to observable graph changes would make the supervision easier to interpret and control. Reliability weighting could emphasize valid, informative outliers while retaining diversity across shift types. Detection performance on previously unseen OOD sources would measure how broadly these learned boundaries apply.

\noindent\textbf{LLMs, GFMs, and GOOD Detection.}
Large language models (LLMs) and graph foundation models (GFMs)
provide semantic judgments, synthetic outliers, and graph--text
scores for GOOD detection~\cite{xu2025llmgood,xu2025graphllm,xu2025glipood}.
Related studies of generalist graph anomaly detection develop
cross-graph transfer and adaptation with limited
labels~\cite{liu2026few,iclr2026owleye,wang2026sagad,qiao2025anomalygfm}.
GOOD detection should become a core capability of GFMs deployed in
open graph environments. Models such as One for All and GraphCLIP
extend graph prediction across domains through shared representations
and pretraining~\cite{liu2024oneforall,zhu2025graphclip}.
Their deployment calls for identifying nodes or graphs outside the
IND scope of the current application. This scope may differ from
the domains and concepts encountered during pretraining: a familiar
concept can be OOD for a downstream task, while a new domain can
be an accepted transfer target. Detection must reflect the
application's rejection criterion across semantic, attribute, and
relational changes.

Future GFMs could take class descriptions, representative nodes or
graphs, and reference substructures as inputs that define the intended
IND population. A detection output alongside task predictions could
combine semantic compatibility with attribute and relational
discrepancy. Pretraining and adaptation could preserve the information
needed for this decision, including unfamiliar combinations of
familiar node content and graph relations. Structural references
could extend detection to whole graphs and graphs with limited text.
Research should examine how prompt changes and model updates affect
detection scores and acceptance thresholds. Selective annotation
could connect rejection to unknown-class discovery and guide updates
to the IND reference~\cite{wen2026oga}. Evaluation should include
both accepted cross-domain inputs and task-defined OOD cases,
measuring detection and downstream prediction separately.

\noindent\textbf{From Score Ranking to Reliable Rejection.}
GOOD detectors should maximize OOD rejection at a specified IND false-rejection budget. CGOD provides a conformal graph-level rejection rule, while adaptive dictionaries update reference evidence at test time~\cite{lin2025cgod,hou2025d2go}. Conformal prediction on graphs and conformal OOD testing with labeled outliers provide statistical starting points for studying relational dependence and auxiliary supervision~\cite{huang2023cfgnn,liang2024integrative}. Future work should establish IND error control for explicit graph sampling schemes and reference-update rules, with detection power assessed against the intended OOD changes. Evaluation should combine controlled interventions with temporal and cross-domain OOD detection, measuring OOD rejection at the IND retention levels required by the application.

\subsection{Conclusion}
\label{sec:conclusion}

This paper presents the first systematic survey dedicated to graph out-of-distribution (GOOD) detection.
We clarify the task and organize 63 core methods into six families according to their primary discrimination mechanisms, examining their information requirements and evaluation settings. Controlled synthetic comparisons deepen the understanding of detection targets and method behavior under different distribution shifts. We also discuss open challenges and research directions to advance reliable GOOD detection.

\bibliographystyle{IEEEtran}
\bibliography{Ref}

\end{document}